\documentclass[review]{elsarticle}

\usepackage{algorithm}
\usepackage{algpseudocode}
\usepackage{subfloat}
\usepackage{subfig}
\usepackage{overpic}
\usepackage{amssymb}
\usepackage{mathtools}
\usepackage{color}
\usepackage{url}

\usepackage{multirow}
\usepackage[figuresright]{rotating}
\floatname{algorithm}{Kernel}

\usepackage{color}

\journal{Journal of \LaTeX\ Templates}

\begin{document}

\begin{frontmatter}

\title{Efficient Parallel Connected Components Labeling with a Coarse-to-fine Strategy}

\author[]{Jun Chen\corref{mycorrespondingauthor}}
\ead{1986ytuak@gamil.com}

\author[]{Keisuke Nonaka}
\ead{ki-nonaka@kddi-research.jp}

\author[]{Hiroshi Sankoh}
\ead{sankoh@kddi-research.jp}

\author[]{Ryosuke Watanabe}
\ead{ru-watanabe@kddi-research.jp}

\author[]{Houari Sabirin}
\ead{ho-sabirin@kddi-research.jp}

\author[]{Sei Naito}
\ead{sei@kddi-research.jp}

\address{Ultra-realistic Communication Group, KDDI Research, Inc. \\
Ohara 2-1-15, Fujimino, Saitama, Japan}



\cortext[mycorrespondingauthor]{Corresponding author}


\begin{abstract}

This paper proposes a new parallel approach to solve connected components on a 2D binary image implemented with CUDA.
We employ the following strategies to accelerate neighborhood exploration after dividing an input image into independent blocks.
In the local labeling stage, a coarse-labeling algorithm, including row-column connection and label-equivalence list unification, is applied first to sort out the mess of an initialized local label map; a refinement algorithm is then introduced to merge separated sub-regions from a single component.
In the block merge stage, we scan the pixels located on the boundary of each block instead of solving the connectivity of all the pixels.
With the proposed method, the length of label-equivalence lists is compressed, and the number of memory accesses is reduced.
Thus, the efficiency of connected components labeling is improved.
Experimental results show that our method outperforms the other approaches between $29\%$ and $80\%$ on average.

\end{abstract}


\end{frontmatter}


\section{Introduction}

Connected components labeling (CCL) is a task to give a unique ID to each connected region in a 2D/3D image.
It means that the input data are clustered as separate groups where the elements from a single group share the same ID.
As a basic data clustering method, CCL is used as a tool for object detection and classification in the field of computer vision and image processing \cite{baxes1994digital} \cite{suzuki2003massive} \cite{suzuki2008mixture}.
W. Song, et al. \cite{song2017motion} presented a motion based skin region of interest detection method using a real-time CCL algorithm to reduce its execution time. 
A fast 3D shape measurement technique using blink-dot projection patterns that utilizes a CCL algorithm to compute the size and location of each dot on the captured images has been reported \cite{chen2013fast} \cite{chen2015blink}.  
P. Guler, et al. proposed a real-time multi-camera video analytics system \cite{guler2016real} employing CCL to perform noise reduction.
Acting as a fundamental operation in all the applications, especially in real-time applications, speeding up CCL is an important task \cite{he2017connected} \cite{cabaret2014review}.

Numerous studies have proposed ways to accelerate CCL.
The proposed solutions on CPU can be summarized into two classes: label propagation algorithms and label-equivalence-based algorithms \cite{he2017connected}.
The approaches \cite{he2011two} \cite{martin2007hybrid} based on label propagation often find an unlabeled pixel using raster scan and give it an unused label; then, the label is propagated to all the pixels in the same region in an irregular way, such as tracing the object's contour \cite{chang2004linear}.  
These approaches are not suitable for parallel implementation and hardware implementation because of the existence of the irregular scan.
The methods \cite{he2009fast} \cite{he2008run} \cite{he2010efficient} \cite{grana2010optimized} \cite{he2014configuration} on the basis of label-equivalence solve the CCL issue with multiple raster scans.
Provisional labels, often associated with the pixel position in the image or in a specific row, are assigned to all the pixels in the first scan; the label-equivalence lists are constructed based on the pixel connectivity and resolved with root-find algorithms in the other steps.  
Since the pixels are processed in a regular way, it is feasible to extend these methods into parallel implementation and hardware implementation \cite{johnston2008fpga} \cite{gu2013fast}.
Until recently, the use of GPUs with interfaces such as CUDA \cite{manohar1989connected} or OpenCL \cite{dewar1987parallel} finds countless applications in both industry and academia areas.
The parallel extension and improvement of serial CCL algorithms are significant advances to enhance the real-time property.
For the algorithms developed on GPUs, data parallelization across multiple processors \cite{nickolls2008scalable} \cite{sanders2010cuda} plays an important role in computing with multiple processing elements in parallel.
Generally, the different data parallelization approaches lead to different computation algorithms.  
According to the various ways of managing data, the reported solutions for CCL on GPUs can be classified into three types: pixel-based algorithms, block-based algorithms, and line-based algorithms.
The first type extends the label-equivalence-like algorithms into parallel ones directly by considering each individual pixel or the pixels in a small group as a computation unit.
The other two types first divide images into independent sections, blocks or lines, then perform local labeling and section merge to solve the CCL.

In this study, we propose a block-based solution to explore the benefit of two-dimensional pixel distribution to reduce the number of iterative operations.
Its main contributions are:
(1), a row-column connection and a label-equivalence list unification algorithm are performed using shared memory to sort out the mess of an initialized local label map;
(2), connectivity analysis is conducted for the pixels on the block boundary instead of all the pixels to reduce the number of memory accesses.
By using our method, the length of the label-equivalence list is compressed and the number of CUDA threads for computation decreases.
In the following sections, we will outline our method, prove the positive effects of coarse-to-fine strategy, and demonstrate its performance.

\section{Previous works}

\subsection{Pixel-based CCL algorithm}

Label-equivalence \cite{hawick2010parallel} is an algorithm that records the lowest label that each label is equivalent to and resolves the equivalence with a small number of iterations.
Jung et al. \cite{jung2010parallel} solved the CCL issue by interactively executing six phases, including initialization, scan, analysis, link, label, and rescan. 
In the scan phase and link phase, they introduce specific masks to construct label-equivalence lists.
In the analysis phase and label phase, they find the roots by tracing each list. 
Kalentov et al. \cite{kalentev2011connected} improved the label-equivalence technique in terms of memory consumption and required processing steps, which removed the reference array and atomic operations in the scan phase.
Soh et al. \cite{soh2014fast} proposed a direction-based searching method that obtains the minimum label by tracing the branches derived from a focused pixel in eight directions.
Block-equivalence \cite{zavalishin2016block} is another extension of label-equivalence solution.
It uses a superpixel block instead of the individual pixel taking into consideration what the pixels located in a $2 \times 2$ block share with the same label with eight-connectivity.
It is effective because the number of candidate pixels for connectivity detection is reduced.
The main drawback of these pixel-based algorithms is that a single label-equivalence list cannot be constructed for one connected component in one scan.
Consequently, the kernels of this algorithm are spawned several times to guarantee that no disjoint equivalence lists exist for a single region.  
Even though some of them reduce the number of iterations at some level, they still need to scan the input image multiple times.
Furthermore, the iterations might vary dramatically in different images.

\subsection{Block-based CCL algorithm}

The parallel version of the union-find algorithm \cite{cormen2009introduction} is presented by Oliveria et al. \cite{oliveira2010study}.
They executed two merges successively, local merge and global merge, to overcome the drawback that it may need to follow a long path to reach the root of two connected pixels.
Although this algorithm outperforms most of the pixel-based CCL algorithms because all the kernels are spawned once, searching for the root of a specific pixel is computationally heavy.
Stava et al. \cite{stava2010connected} designed a solution in the similar manner.
In the local merge stage, they improved the label-equivalence algorithm by implementing all iterations inside the kernel, such that no synchronization between host and device is required.
In the global merge stage, they use the connectivity between all border elements of two neighboring blocks to merge their equivalence lists.
It is necessary to perform the operations of global merge several times to guarantee that all equivalence lists are merged.
Kumar et al. \cite{kumar2016real} implemented the CCL algorithm using a divide and conquer technique \cite{park2000fast} on CUDA that solves the local connection using the Floyd-Warshall algorithm \cite{cormen1990floyd} and merges blocks by considering three different cases. 
Here, the various processing approaches for the three cases leads to thread divergence thus limiting the performance.

\subsection{Line-based CCL algorithm}

Chen et al. \cite{chen2011block} proposed a two-scan approach, extended from a stripe-based CCL method \cite{zhao2010stripe}, to process stripe extraction and stripe union, respectively.
The first scan can run in parallel by using shared memory, while the second scan is a sequential operation.
ACCL \cite{paravecino2014accelerated} is another parallelization algorithm that decomposes the image into rows.
By defining a span as a group of pixels that are located contiguously in a row with the same intensity, it spawns two kernels, find spans and merge spans, to label an input image.
The involvement of dynamic parallelism means that good performance can be achieved with this method.
However, it is not suitable to process large images because there is a limitation on the number of threads in one block \cite{xu2014graph}.
Yonehara et al. proposed a line-based solution \cite{yonehara2015line} that improves the local labeling phase of the conventional union-find algorithm \cite{oliveira2010study} by conducting a row unification using shared memory.
The absence of merge algorithm makes it label an individual section efficiently in the first scan, while the other scans cannot make any further improvements.

\begin{figure}[t]
\centering
\footnotesize
	\begin{minipage}[b]{0.48\linewidth}
 		\centering
 		\subfloat[Input data]
		{
 	 		\begin{overpic}[width=1\textwidth]
 	 			{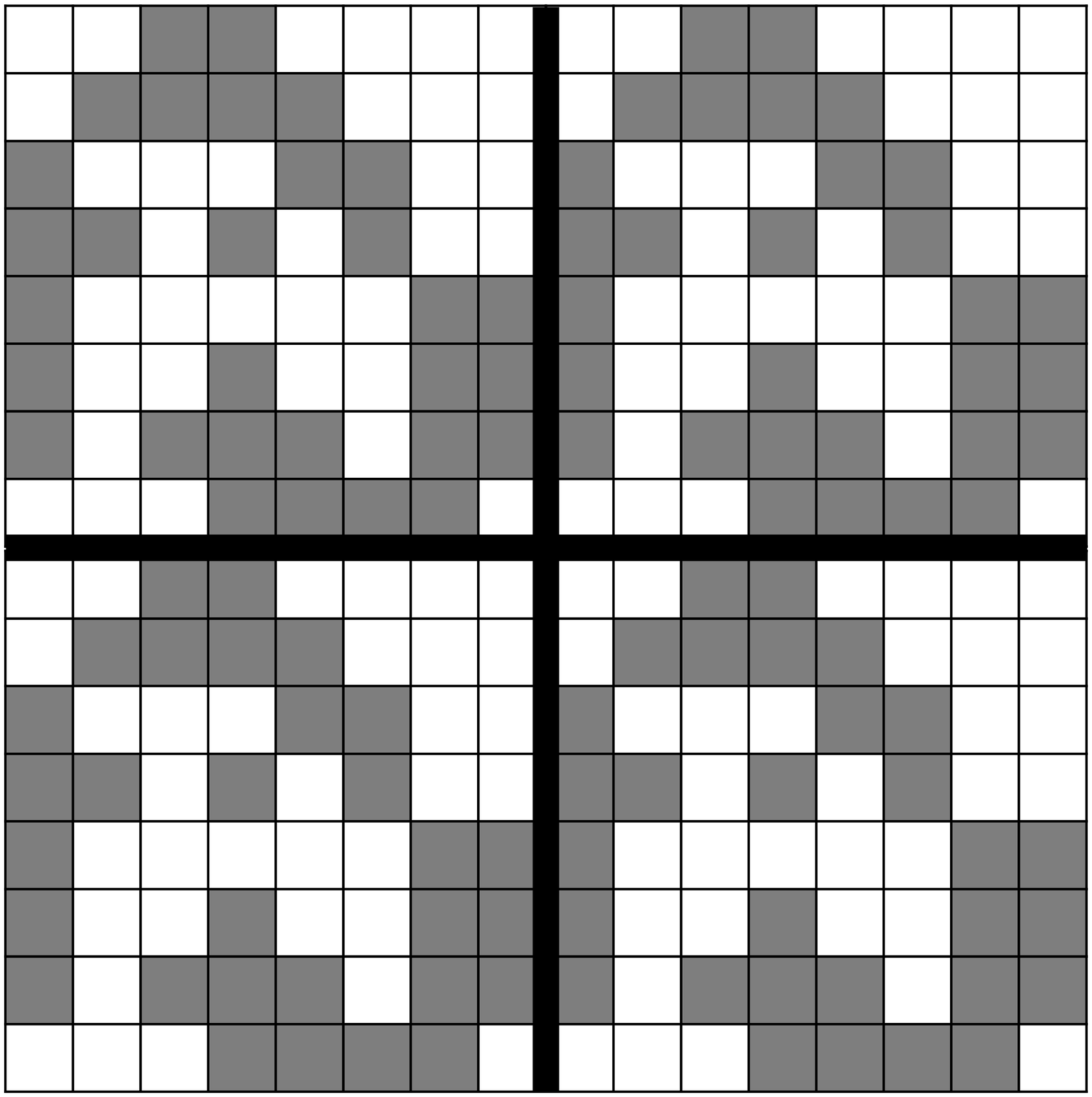}
 		\end{overpic}
 	 	}
	\end{minipage}
\hskip 2mm
	\begin{minipage}[b]{0.48\linewidth}
 		\centering
 		\subfloat[Initialized local label map]
		{
 			\begin{overpic}[width=1\textwidth]
 	 			{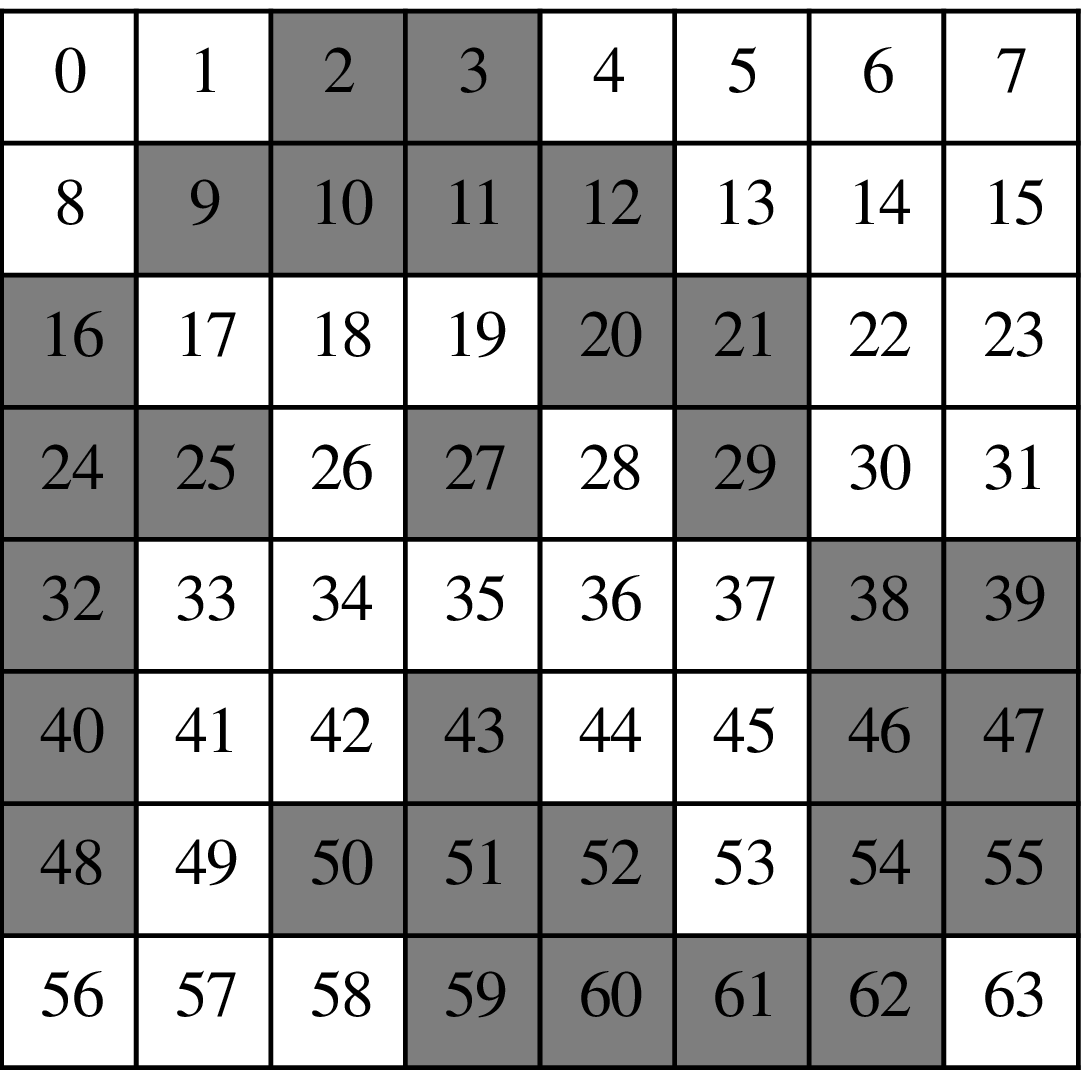}
 		\end{overpic}
 		}
	\end{minipage}
\caption{Input data and initialized local label map.}
\label{fig:initialization}
\vskip -3mm
\end{figure}

\begin{figure}[t]
\centering
\footnotesize
	\begin{minipage}[b]{0.48\linewidth}
 		\centering
 		\subfloat[Label-equivalence list after row scan]
		{
 	 		\begin{overpic}[width=1\textwidth]
 	 			{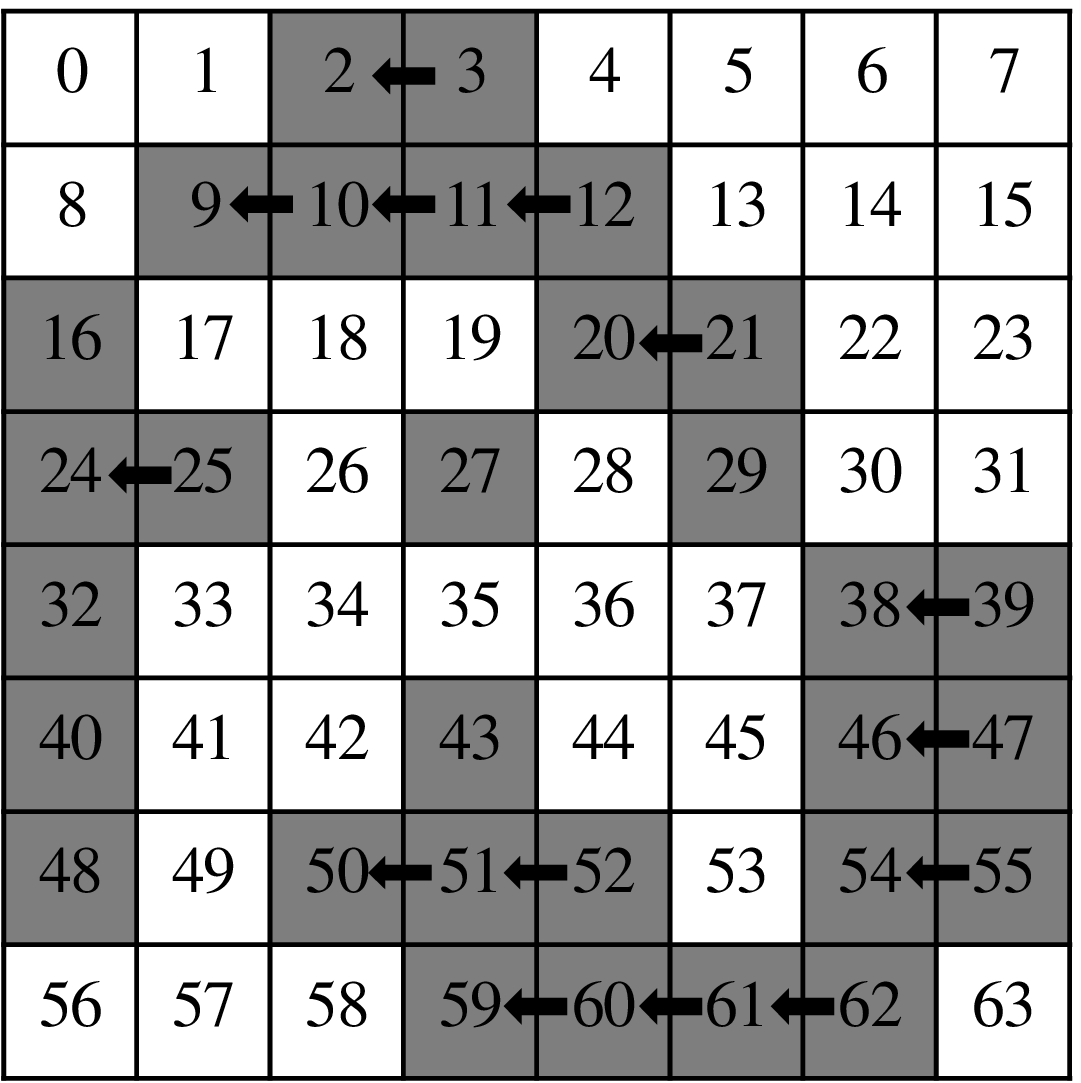}
 		\end{overpic}
 	 	}
	\end{minipage}
\hskip 2mm
	\begin{minipage}[b]{0.48\linewidth}
 		\centering
 		\subfloat[Label-equivalence list after column scan]
		{
 			\begin{overpic}[width=1\textwidth]
 	 			{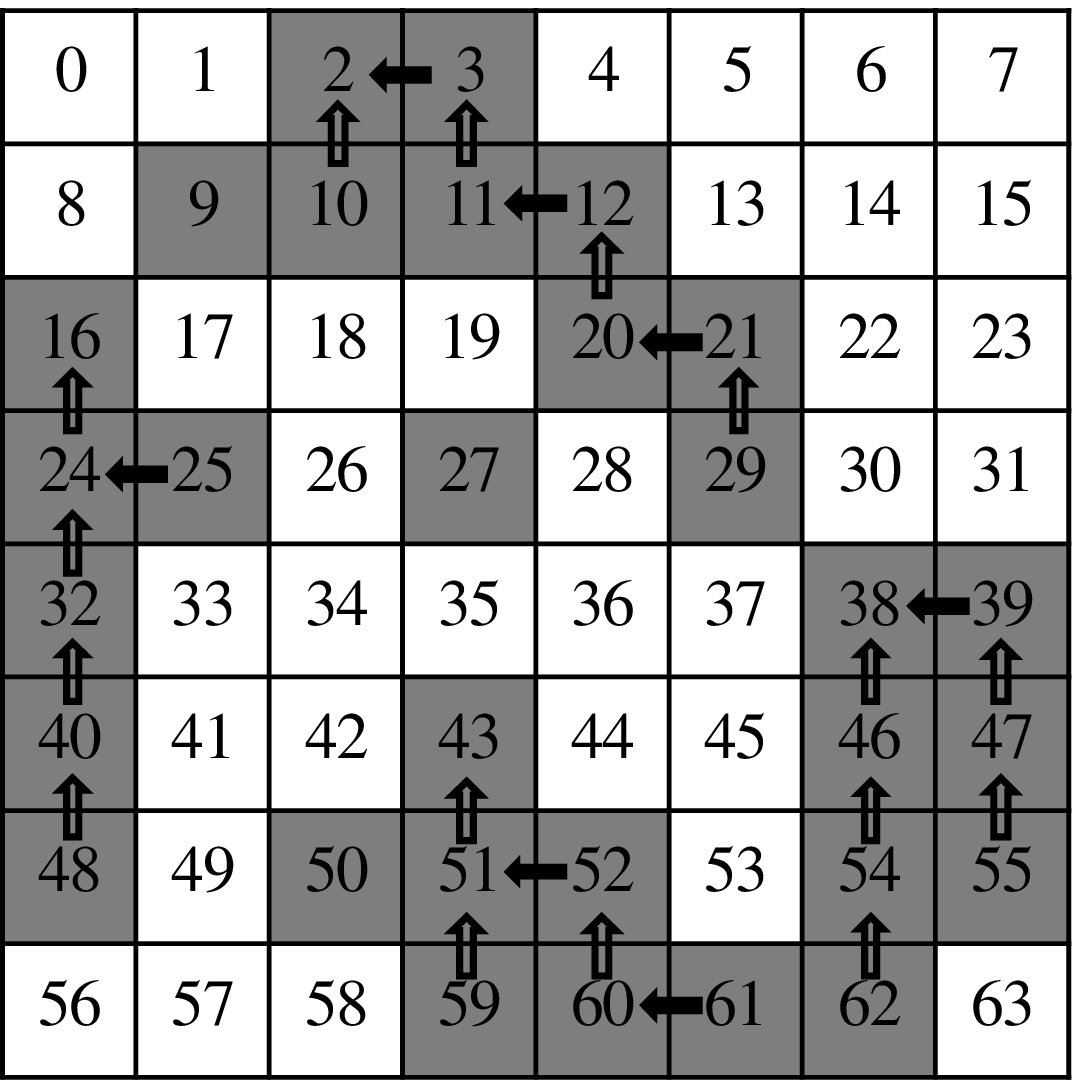}
 		\end{overpic}
 		}
	\end{minipage}
\caption{Coarse label-equivalence construction.}
\label{fig:coarse-label-equivalence-construction}
\vskip -3mm
\end{figure}

\section{Algorithm and implementation}

We assume that a pixel in an image has three attributes comprising position, intensity, and label.
Position means its raster scan order in row-major order, which can be expressed by $P(x,y) = x + y * W$. 
Here $(x,y)$ is  its 2D coordinate in the image.
$(H, W)$ is the resolution of the image plane.
Intensity is the color intensity of the pixel, which can be expressed by $I(x,y)$.
In our implementation, $I(x,y) = 1$ when a pixel belongs to foreground, and $I(x,y) = 0$ when it is background.
Label $L(x,y)$ is what we should find to depict each connected region.

There are three steps in our method to solve the segmentation of an image. 
In the first step, we divide the input image into blocks, and perform local labeling using a coarse-to-fine strategy.
In the second step, we extract the pixels located on the boundary of each block to be a sub-image, and construct global label-equivalence lists.
In the last step, the final label map is obtained by solving equivalence using a root-find algorithm.

\begin{algorithm}[!htp]
	\caption{\textbf{Local labeling with coarse-to-fine strategy}}
	\label{Alg:LocalMerge}
	\begin{algorithmic}
		\Require $label_{sm}[],subimg_{sm}[]$\text{ are on shared memory}        
		\Require $labelmap[]$ \text{is on global memory}

		\State \textbf{declare} \text{int} $x,y,tid,temp,l,l_x,l_y,g_l,label_{sm}[],dBuff_{sm}[]$
		
		\State $x,y \gets$ \text{2D global thread id}			
		\State  $tid \gets$ \text{1D thread id within block}		
		\State  $label_{sm}[tid] \gets tid$  
		\State  $subimg_{sm}[tid] \gets image[x,y]$  
		\State  \text{call syncthreads()}
		
		\State  \text{// row scan}
		\State  $\textbf{if} \ \ subimg_{sm}[tid] == subimg_{sm}[tid-1]$ 
		\State  \ \ \ \ $label_{sm}[tid] = label_{sm}[tid-1]$
		\State  $\textbf{end if}$
		\State  \text{call syncthreads()}
		
		\State  \text{// column scan}
		\State  $\textbf{if} \ \ subimg_{sm}[tid] == subimg_{sm}[tid-blockdim.x]$ 
		\State  \ \ \ \ $label_{sm}[tid] \gets label_{sm}[tid--blockdim.x]$
		\State $\textbf{end if}$
		\State  \text{call syncthreads()}

		\State  $label_{sm}[tid] \gets findroot(label_{sm}[], tid)$
		
		\State  \text{// refinement (row scan)}
		\State  $\textbf{if} \ \ subimg_{sm}[tid] == subimg_{sm}[tid-1]$ 
		\State  \ \ \ \ \ $merge(label_{sm}[], tid, tid - 1)$
		\State  $\textbf{end if}$
		\State  \text{call syncthreads()}	
				
		\State  $l \gets findroot(label_{sm}[], tid)$
		
		\State  \text{// convert local index to global index}		
		\State  $(l_x, l_y)\gets (l \ / \  blockdim.x, l \ \% \ blockdim.x)$
		\State  $g_l \gets (blockIdx.x * blockDim.x + l_x) + (blockIdx.y * blockDim.y + l_y) * imgwidth$
		\State  $labelmap[x,y] \gets g_l$	
		
	\end{algorithmic}
\end{algorithm}

\subsection{Local labeling with coarse-to-fine strategy}

The first step, local labeling with a coarse-to-fine strategy, consists of four phases: initialization, coarse labeling, refinement, and ID conversion.

\begin{figure}[t]
\footnotesize
\centering
\includegraphics[width=0.48\linewidth]{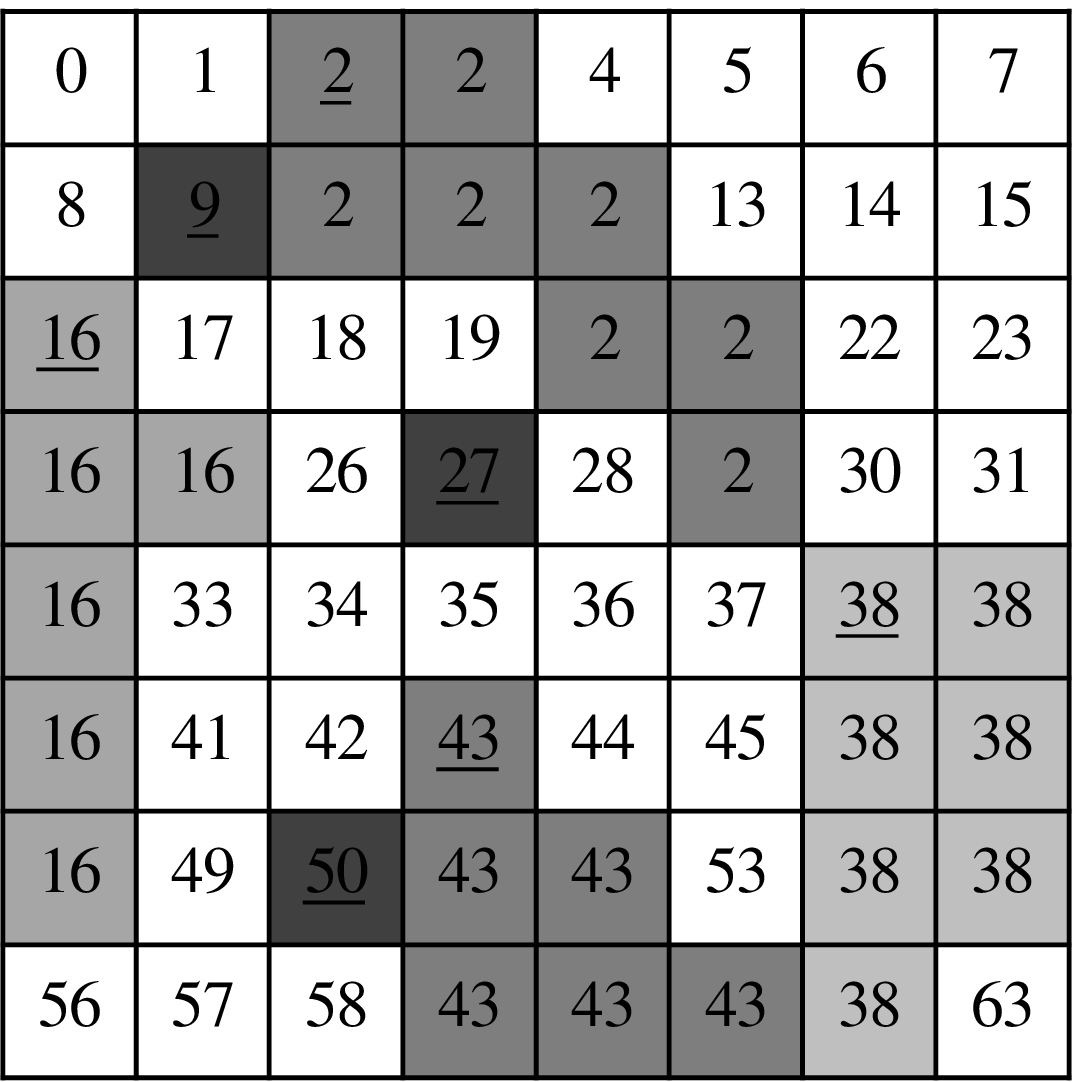}
\caption{Label-equivalence list after row-column connection.}
\label{fig:solve-coarse-label-equivalence-list}
\end{figure}%

\begin{figure}[t]
\centering
\footnotesize
	\begin{minipage}[b]{0.31\linewidth}
 		\centering
 		\subfloat[A single region]
		{
 	 		\begin{overpic}[width=1\textwidth]
 	 			{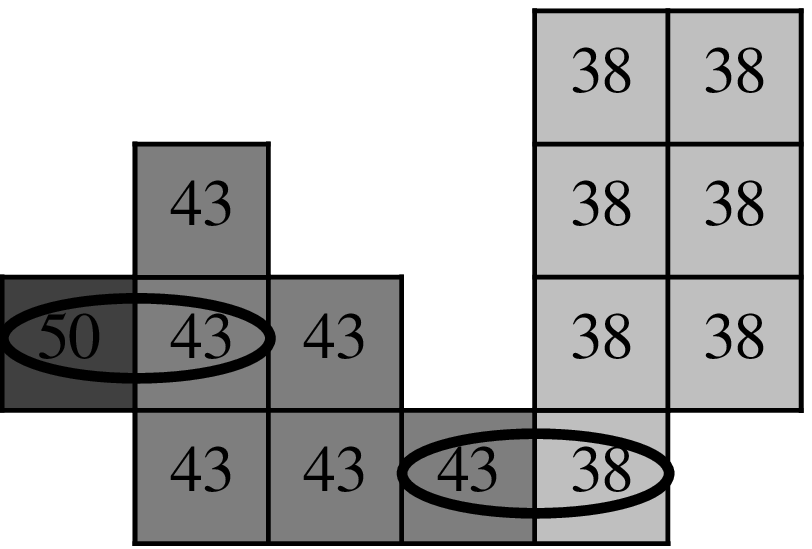}
 		\end{overpic}
 	 	}
	\end{minipage}
\hskip 2mm
	\begin{minipage}[b]{0.31\linewidth}
 		\centering
 		\subfloat[Refined list]
		{
 			\begin{overpic}[width=1\textwidth]
 	 			{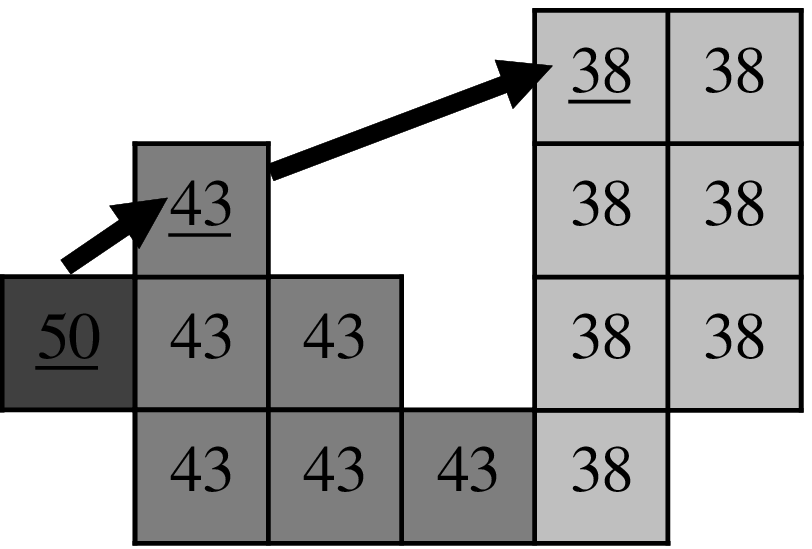}
 		\end{overpic}
 		}
	\end{minipage}
\hskip 2mm
	\begin{minipage}[b]{0.31\linewidth}
 		\centering
 		\subfloat[Refined local label map]
		{
 			\begin{overpic}[width=1\textwidth]
 	 			{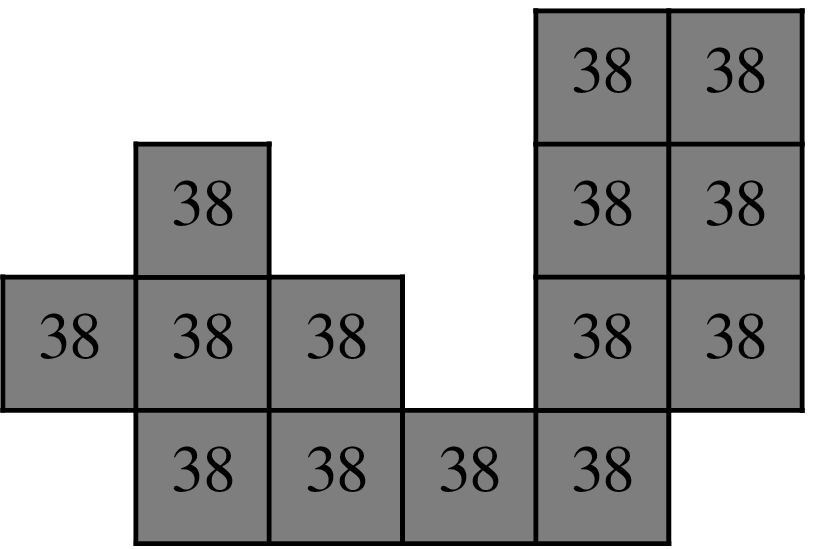}
 		\end{overpic}
 		}
	\end{minipage}	
\caption{Refine local label map.}
\label{fig:refine-local-label-map}
\vskip -3mm
\end{figure}

\subsubsection{Initialization}

In our algorithm, each pixel should be assigned with a provisional label first so that a connection-list can be constructed and solved.
We use $L(x, y) = P(x, y)$ for indicating the provisional label of $(x,y)$.
In this way, the root of an equivalence list is the element with the label equal to the address of the element itself.
Fig.~\ref{fig:initialization} gives an illustration of this step where (a) shows that a $16 \times 16$ binary image is split into four sub-images with a resolution of $(H, W) = (8, 8)$; (b) presents an example of an initialized local label map.
The grey elements on the image, $I(x, y) = 1$, express foreground pixels.
In our CUDA implementation, we dispatch the sub-images to various GPU threads blocks where the threads can cooperate with each other using shared memory and can be synchronized \cite{nvidia2015toolkit}.
The provisional labels and pixel positions are associated with the thread ID within a threads block as is shown in the steps $1-6$ of Algorithm~\ref{Alg:LocalMerge}.
Threads synchronization is necessary because the buffer could not be initialized by multiple threads at the same time. 

\subsubsection{Coarse labeling}

In an initialized local label map, the provisional label of the left pixel and that of the upper pixel are always smaller than the label of a target pixel, while the upper one is always the minimum.
Based on this fact, we scan row and column successively to make a coarse label-equivalence list.
In the case of a row scan, we associate two consecutive pixels by updating the label of the right pixel with the left label $L(x,y) = L(x-1,y)$ if both of them are foreground $I(x,y) = I(x-1,y) = 1$.
Label-equivalence trees are constructed for continuous foreground pixels of each row with the method presented in Fig.~\ref{fig:coarse-label-equivalence-construction} (a).
The scanning approach along the vertical direction is performed in the same manner where the association between left and right is updated by the association between up and down if all of the three pixels are foreground. 
Fig.~\ref{fig:coarse-label-equivalence-construction} (b) presents a demonstration of coarse label-equivalence lists after column scan.
The same result can be achieved by comparing the labels of the above-mentioned three pixels directly at the same time.
However, we find the proposed method is faster because it does not involve branch divergence and the boundary-related operations.
Unlike the methods that records the entire equivalence, this method records the lowest neighbor label that the label is equivalent to.
Its memory access complexity is reduced due to the utilization of shared memory, while the list can be unified by a low number of iterations.
Fig.~\ref{fig:solve-coarse-label-equivalence-list} illustrates the segments of a coarse local label map and the root of each list.
It was found that this step can not provide a complete segmentation but splits a connected region into several groups.
The pseudo code for this step is listed in steps $8-17$ of Algorithm~\ref{Alg:LocalMerge}.

\begin{figure}[t]
\footnotesize
\centering
\includegraphics[width=0.97\linewidth]{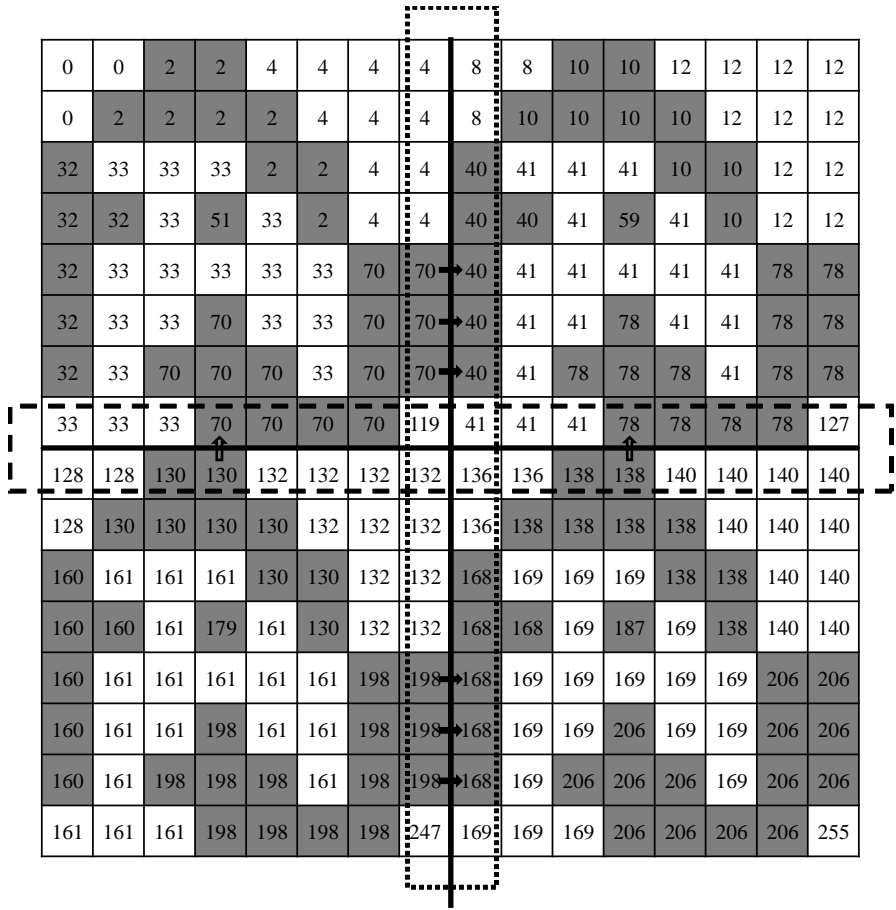}
\caption{Label map after local labeling.}
\label{fig:block-merge-with-boundary-analysis}
\end{figure}%

\begin{figure}[t]
\centering
\footnotesize
	\begin{minipage}[b]{0.48\linewidth}
 		\centering
 		\subfloat[A connected region]
		{
 	 		\begin{overpic}[width=1\textwidth]
 	 			{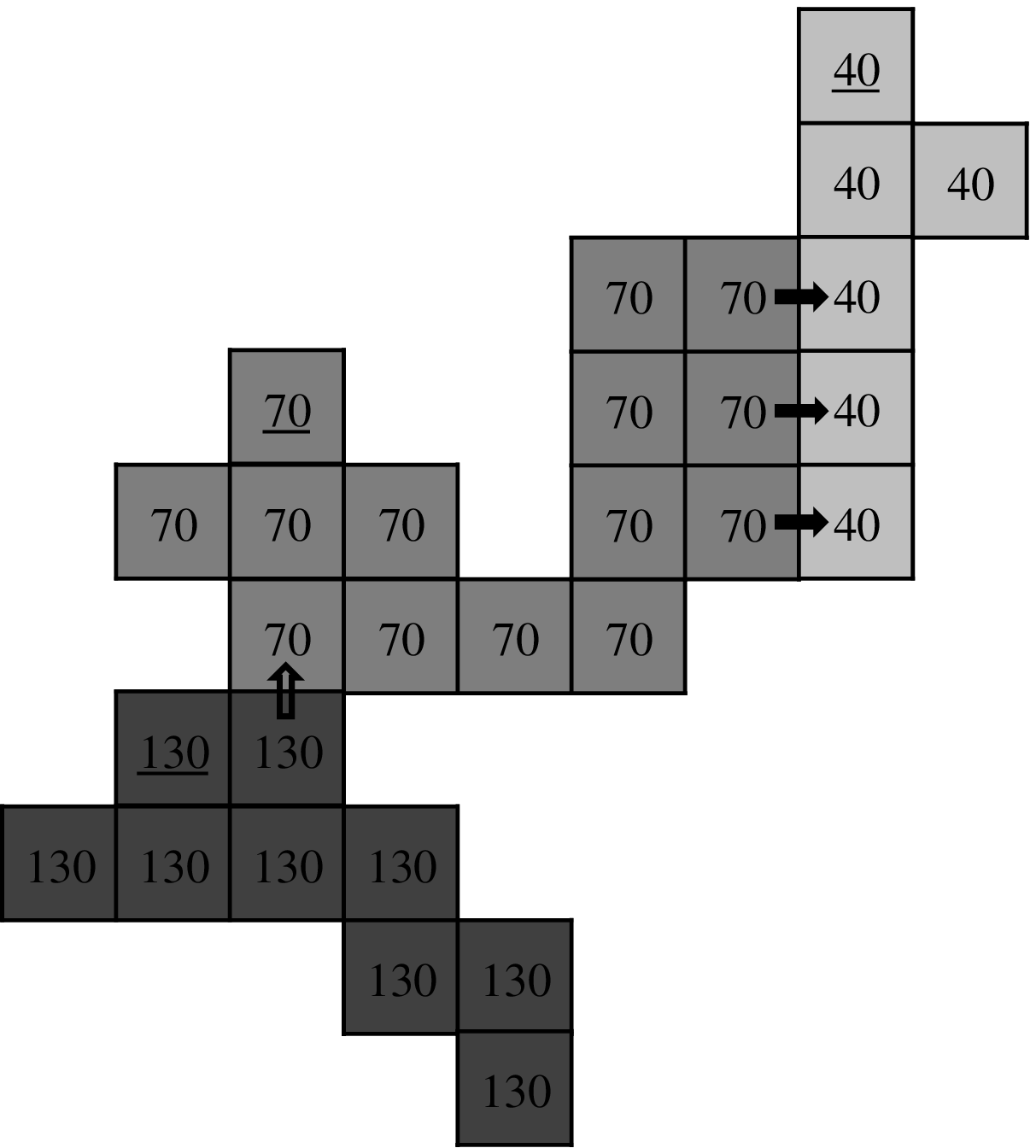}
 		\end{overpic}
 	 	}
	\end{minipage}
\hskip 2mm
	\begin{minipage}[b]{0.48\linewidth}
 		\centering
 		\subfloat[Global label-equivalence list]
		{
 			\begin{overpic}[width=1\textwidth]
 	 			{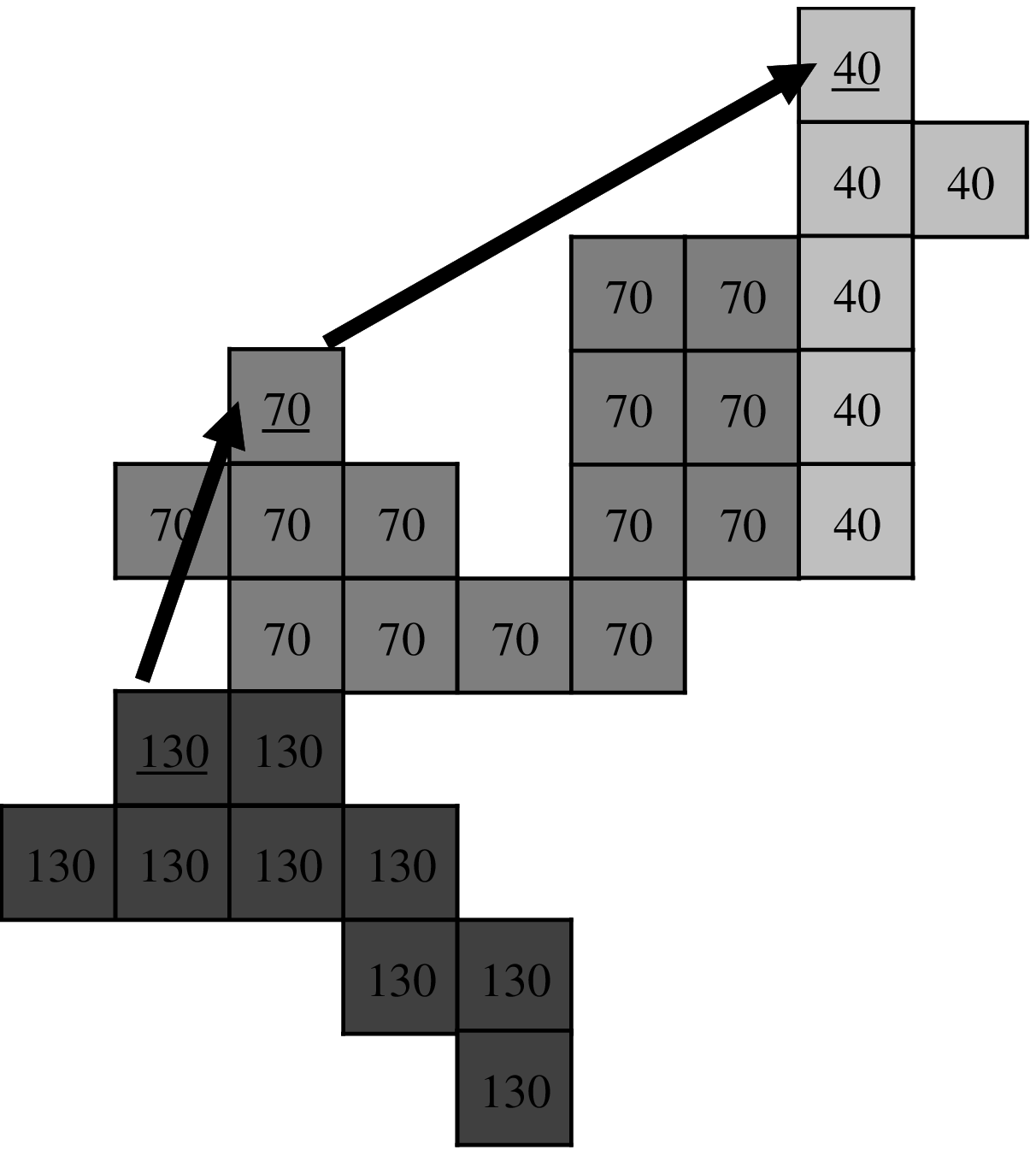}
 		\end{overpic}
 		}
	\end{minipage}
\caption{Global label-equivalence construction.}
\label{fig:global-label-equivalence-construction}
\vskip -3mm
\end{figure}

\subsubsection{Refinement}

This phase is a task to merge the segments that belong to a single region.
As shown in Fig.~\ref{fig:refine-local-label-map} (a), three isolated sub-regions exist in the one connected region.
The pixels in the ellipse are the branch dividing points that lead pixels to different label-equivalence lists.
Corresponding to the initialization approach of the provisional label, the branch dividing points are always in the horizontal direction.
So the sub-regions can be merged together with another row scan.
If two consecutive foreground pixels have different labels, we compare their roots and make the larger one point to the smaller one.
It should be noticed that the atomic operation is necessary here because the same equivalence list may be updated by multiple threads at the same time.
With this method, a new label-equivalence list is available as illustrated in the example in Fig.~\ref{fig:refine-local-label-map} (b).
Finally, the region is unified using a root-find algorithm as shown in Fig.~\ref{fig:refine-local-label-map} (c). 
The pseudo code for refinement is listed in steps $19-23$ of Algorithm~\ref{Alg:LocalMerge}.

\subsubsection{ID conversion}

The final step of local labeling is an ID conversion that converts the local index to global index and transfers the result to global memory.
The global index identifies the $1-$D address of a pixel in the entire image.
Steps $25-26$ of Algorithm~\ref{Alg:LocalMerge} give its pseudo code.

\begin{algorithm}[!ht]
	\caption{\textbf{Boundary analysis}}
	\label{Alg:blockmerge}
	\begin{algorithmic}
		\Require \text{both block dimension and grid dimension are 2D}		
		\Require $labelmap[]$ \text{is on global memory}
		
		\State \textbf{declare} \text{int} $id,h_x,h_y,v_x,v_y, pInLine, p_h, p_v$
		\State \textbf{declare} \text{bool} $b_h, b_v$
		\State $id \gets$ \text{1D global thread id}
		
		\State \text{// convert 1D global index id to 2D image index}
		\State $h_x \gets id \ \% \ imgwidth$
		\State $h_y \gets id \ / \ (imgwidth * blockdim.y)$
		
		\State $pInLine \gets imgwidth \ / \ blockDim.x$
		\State $v_x \gets id \ \% \ pInLine * blockDim.x$
		\State $v_y \gets id \ / \ pInLine$	
		
		\State \text{// boundary analysis along x-axis}
		\State $\textbf{if} \ \  image[h_x,h_y] == image[h_x-1,h_y]$
		\State  \ \ \ \ $merge(labelmap, \ p_h, \ p_h - 1)$;
		\State $\textbf{end if}$	
		
		\State \text{// boundary analysis along y-axis}
		\State $\textbf{if} \ \  image[v_x,v_y] == image[v_x,v_y-imgwidth]$
		\State  \ \ \ \ $merge(labelmap, \ p_v, \ p_v - imgwidth)$;		
		\State $\textbf{end if}$
	\end{algorithmic}
\end{algorithm}

\subsection{Block merge with boundary analysis}

In the block merge phase, we perform connective detection for the pixels on the block boundary to merge the equivalence lists of the same connected component from different blocks.
Assuming the resolution of an input image is $N \times M$ and the block configuration of Kernel 1 is $\left\{b_x, b_y, 1\right\}$, the number of border pixels along the $x-$axis $P_x$ and the number of border pixels along the $y-$axis $P_y$ can be determined as follows:
\begin{eqnarray}
P_x = \left \lfloor N \ / \ b_x\right \rfloor  * M, \\
P_y = \left \lfloor M \ / \ b_y\right \rfloor  * N,
\end{eqnarray}
Here, $\left \lfloor x \right \rfloor$ means the largest integer smaller or equal to $x$. 
It is found that the candidate pixels for detection get shrunk by $b_x$ times for boundary analysis along the $x-$axis and $b_y$ times for boundary analysis along the $y-$axis.
Similar to coarse labeling, we scan the vertical boundary and the horizontal boundary successively.
If two consecutive foreground pixels, vertical or horizontal, are foreground, we link them with the larger label pointing to the smaller one.
Fig.~\ref{fig:block-merge-with-boundary-analysis} shows a global label map after local labeling.
The pixels in rectangles are border pixels.
The arrows show the association between two pixels.
Fig.~\ref{fig:global-label-equivalence-construction} (a) presents a connected region that is composed of three sub-regions from three blocks.
Fig.~\ref{fig:global-label-equivalence-construction} (b) gives a global label-equivalence list by connecting the root of each sub-region.
In our implementation, $\max\{P_x,P_y\}$ threads should be invoked to integrate the boundary analysis along the $x-$ and $y-$axes into one kernel.  
Steps $1-9$ of Algorithm~\ref{Alg:blockmerge} illustrates how to compute the global index of a border pixel, while steps $10-16$ indicate the merge operation.

\subsection{Update global label map}

When all blocks are merged, the independent local label maps are associated as an entirety.
The final global label map represents the complete segmentation of an input image where every equivalence list corresponds to an unbroken connected component.
The roots of global label-equivalence lists can be obtained by using the root-find algorithm that is the same as what was used in Algorithm~\ref{Alg:LocalMerge}.

%
%
%

\section{Comparative evaluation}

In order to demonstrate the performance of our method, we compare it with the following approaches.

\begin{itemize}
\item[-] C2FL as our proposed method.
\item[-] RC2FL as a revision of our method with coarse labeling only along row.
\item[-] CC2FL as a revision of our method with coarse labeling only along column.
\item[-] NC2FL as a revision of our method without coarse-to-fine strategy.
\item[-] LE \cite{kalentev2011connected} as a conventional pixel-based label equivalence solution.
\item[-] BE \cite{zavalishin2016block} as a more recent representative of the pixel-based solution using block equivalence technique.
\item[-] SMCCL \cite{stava2010connected} as a more recent block-based method using shared memory.
\item[-] UF \cite{oliveira2010study} as a more recent block-based method with a union-find method. 
\item[-] LUF \cite{yonehara2015line} as a more recent and fast representative of the line-based method. 
\item[-] CPUCCL \cite{grana2016optimized} as a more recent implementation on CPU.
\end{itemize}

There are two kinds of comparative experiments.
The first is an effectiveness evaluation of coarse labeling which compares C2FL with RC2FL, CC2FL and NC2FL.
It should be noted that the local labeling works correctly if refinements along both row and column are applied even without coarse labeling.
When each individual pixel is considered as a sub-region of a connected region, the third phase, refinement, is able to generate an entire local label-equivalence list.  
The significance of coarse labeling is that it allows to local merge to be performed efficiently.
The second is a comparison with existing CUDA-based algorithms and a latest CPU-based sequential algorithm.
The execution times of C2FL, RC2FL, CC2FL, NC2FL, LE, BE, SMCCL, UF, LUF, and CPUCCL for datasets \cite{grana2016yacclab} are listed.

All the experiments were performed on a PC Intel(R) Core(TM) i7-6700K CPU, 4.00 GHz $\And$ 4.00 GHz, 32.0 GB RAM, NVIDIA Geforce GTX 1070 with Windows 7 Professional Service Pack 1.
All the algorithms were implemented in C++ language by use of OpenCV 2.4.13 and CUDA 8.0.

\begin{figure}[t]
\footnotesize
\centering
\includegraphics[width=0.90\linewidth]{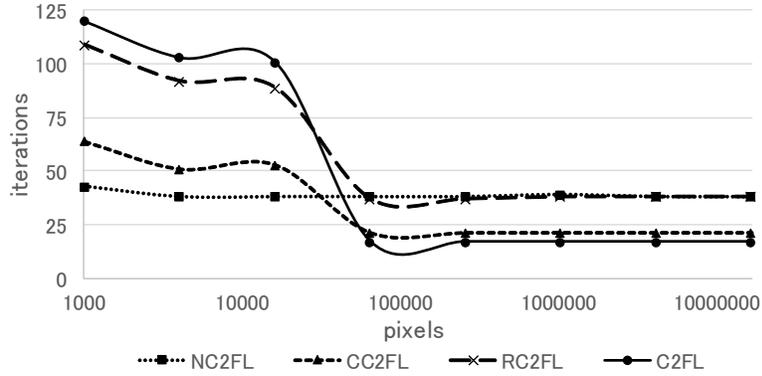}
\vskip 1mm
(a) The number of iteration versus the image size
\vskip 2mm

\includegraphics[width=0.90\linewidth]{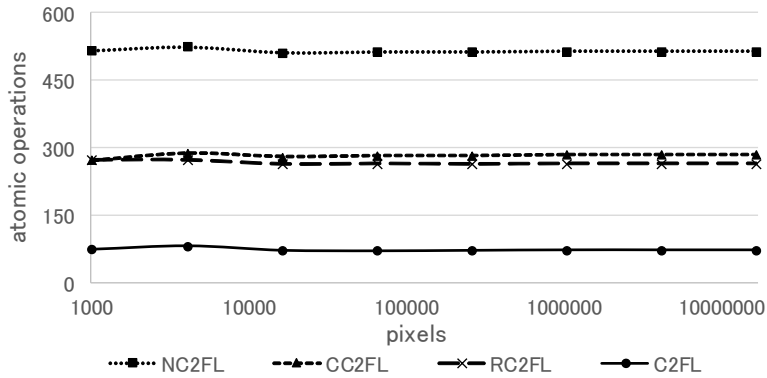}
\vskip 1mm
(b) The number of atomic operation versus the image size
\caption{The number of iteration and atomic operation versus image size}
\label{fig:iteration-and-atomic-operation-verssus-image-size}
\end{figure}%

\begin{figure}[t]
\footnotesize
\centering
\includegraphics[width=0.90\linewidth]{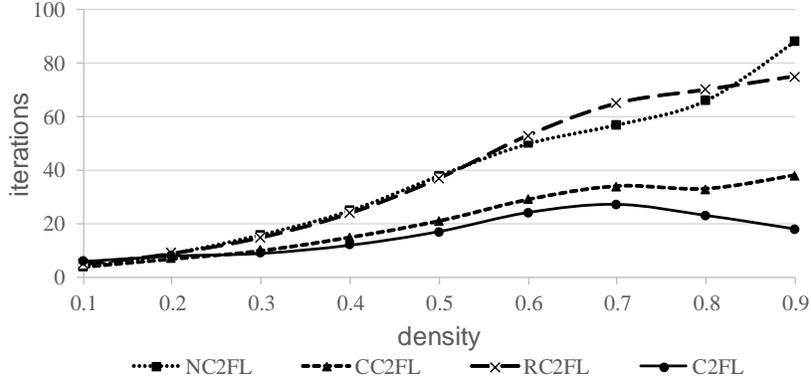}
\vskip 1mm
(a) The number of iteration versus noise density
\vskip 2mm

\includegraphics[width=0.90\linewidth]{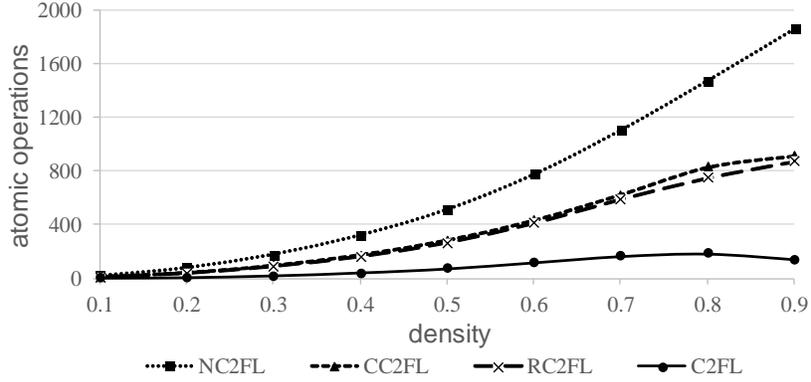}
\vskip 1mm
(b) The number of atomic operation versus noise density
\caption{The number of iteration and atomic operation versus noise density}
\label{fig:iteration-and-atomic-operation-versus-noise-density}
\end{figure}%

\begin{figure}[t]
\centering
\footnotesize

	\begin{minipage}[b]{0.40\linewidth}
 		\centering
 		\subfloat[C2FL]
		{
 	 		\begin{overpic}[width=1\textwidth]
 	 			{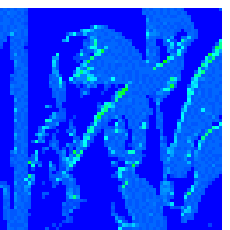}
 		\end{overpic}
 	 	}
	\end{minipage}
\hskip 3mm
	\begin{minipage}[b]{0.40\linewidth}
 		\centering
 		\subfloat[RC2FL]
		{
 			\begin{overpic}[width=1\textwidth]
 	 			{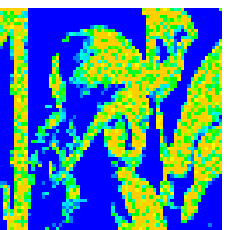}
 		\end{overpic}
 		}
	\end{minipage}
\vskip 1.5mm
	\begin{minipage}[b]{0.40\linewidth}
 		\centering
 		\subfloat[CC2FL]
		{
 			\begin{overpic}[width=1\textwidth]
 	 			{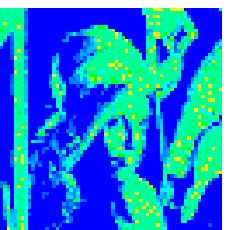}
 		\end{overpic}
 		}
	\end{minipage}	
\hskip 3mm
	\begin{minipage}[b]{0.40\linewidth}
 		\centering
 		\subfloat[NC2FL]
		{
 			\begin{overpic}[width=1\textwidth]
 	 			{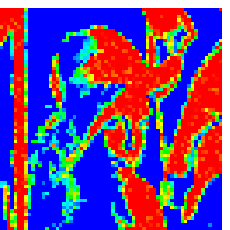}
 		\end{overpic}
 		}
	\end{minipage}	

\caption{Color map of the number of iterations.}
\label{fig:color-map-of-number-of-iterations}
\vskip -3mm
\end{figure}

\begin{figure}[t]
\centering
\footnotesize

	\begin{minipage}[b]{0.40\linewidth}
 		\centering
 		\subfloat[C2FL]
		{
 	 		\begin{overpic}[width=1\textwidth]
 	 			{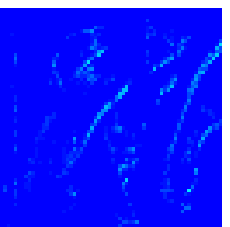}
 		\end{overpic}
 	 	}
	\end{minipage}
\hskip 3mm
	\begin{minipage}[b]{0.40\linewidth}
 		\centering
 		\subfloat[RC2FL]
		{
 			\begin{overpic}[width=1\textwidth]
 	 			{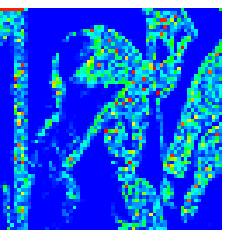}
 		\end{overpic}
 		}
	\end{minipage}
\vskip 1.5mm
	\begin{minipage}[b]{0.40\linewidth}
 		\centering
 		\subfloat[CC2FL]
		{
 			\begin{overpic}[width=1\textwidth]
 	 			{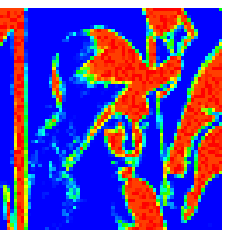}
 		\end{overpic}
 		}
	\end{minipage}	
\hskip 3mm
	\begin{minipage}[b]{0.40\linewidth}
 		\centering
 		\subfloat[NC2FL]
		{
 			\begin{overpic}[width=1\textwidth]
 	 			{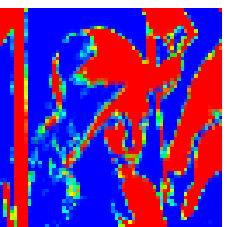}
 		\end{overpic}
 		}
	\end{minipage}		

\caption{Color map of the number of atomic operations.}
\label{fig:color-map-of-number-of-atomic-operations}
\vskip -3mm
\end{figure}

\subsection{Effectiveness of coarse labeling}

There are two significant factors, the number of iterations and the number of atomic operations, in one threads block affect the efficiency of local labeling seriously.
Iteration refers to the process of iterating an operation such as tracing a label-equivalence list to find its root.
Most of the algorithms that are expressed in C++ language take only a few of lines. 
However, there may be thousands of instructions that are executed on hardware.
Generally, the number of iterations reflects the number of instructions and the program$'$s execution time.
Atomic operations are a kind of processing performed without interference from any other threads.
They are often essential for multithreaded applications to prevent race conditions, especially when different threads attempt to modify and write the same memory address.
If two or more threads perform an atomic operation at the same memory address at the same time, those operations will be serialized.
This means that the more atomic operations the slower the execution.  

In our evaluation, the threads block of CUDA was configured as $\left\{ 32, 32, 1\right\}$. 
We use a set of square binary images with various resolution and random noise to show the difference in the number of iterations as well as the number of atomic operations among C2FL, RC2FL, CC2FL, and NC2FL.
There are nine different foreground densities, from $0.1$ to $0.9$, and eight resolutions, from a low resolution of $32 \times 32$ pixels to a maximum resolution of $4096 \times 4096$ pixels.
The experiments provide us an opportunity to evaluate the performance of coarse labeling both in terms of scalability on the number of pixels and in terms of scalability on the density of connected regions.

Fig.~\ref{fig:iteration-and-atomic-operation-verssus-image-size} shows how the the number of iterations and atomic operations of different algorithms change with images of increasing size.
Here, the label density is $0.5$, which remains in all the images.
The reported results were the average of all the launched threads blocks of 100 runs of each algorithm.
As presented in (a) of Fig.~\ref{fig:iteration-and-atomic-operation-verssus-image-size}, it can be seen that the iteration for coarse-labeling, regardless of whether it is full coarse-labeling or partial coarse-labeling, are heavier than those of NC2FL when the number of pixels is less than $65535$ ($256 \times 256$ image),
while these iterations equal to or less than those of NC2FL when the pixels of an image exceed $65535$.
The phenomenon is reasonable because a local label-equivalence list in a low-resolution image is short.
Under the circumstances, the compression by coarse labeling can not reduce but increase the number of iterations. 
The linear independence of the number of atomic operations with respect to the image size can be observed in (b) of Fig.~\ref{fig:iteration-and-atomic-operation-verssus-image-size}.
The proposed algorithm always takes the fewest atomic operations to segment an input image.

Fig.~\ref{fig:iteration-and-atomic-operation-versus-noise-density} highlights the behavior of the algorithms varying the foreground label densities of a $2048 \times 2048$ image.
It can be proved that both of the factors have a significant linear correlation with the label densities, and our method has the best performance among all the densities.
This result is logical because coarse labeling reduces complexity and facilitates the solution. 

The third experiment demonstrates the efficiency of coarse labeling using a binary Lena image with a size of $2048 \times 2048$.
Fig.~\ref{fig:color-map-of-number-of-iterations} shows the color map of the maximum iterations of each threads block of each algorithm where the darker color expresses more numerous iterations.
It indicates that our proposed method solves the CCL issue with the fewest iterations.
Fig.~\ref{fig:color-map-of-number-of-atomic-operations} expresses the number of atomic operations in the same manner.
With regard of RC2FL, CC2FL, and NC2FL, we find that most of the race conditions occur on the blocks holding pixels from a flat foreground region.
Nevertheless, there is no risk of a race condition for the blocks in C2FL.
Meanwhile, our approach evidences that the minimum atomic operations are required for a block with an edge area.

\begin{sidewaystable}[htbp]
\centering

    \caption{Execution time in millisecond for different images}     
    \label{tab:speedup}
	\centering

    \begin{small}
	\begin{tabular}{ |c|c|c|c|c|c|c|c|c|c|c|c| }
	\hline
	Images &  & {CPUCCL} & {LE} & {BE}& {CCLSM} & {UF} & {LUF} & {NC2FL} & {RC2FL} & {CC2FL} &{C2FL} \\ 
	\hline
	\multirow{3}{*}
	{\!\!\!\!\! lena \!\!\! $(\! 512 \! \times \! 512 \!)$ \!\!\!\!\!} 
 	& min & 0.371	& 0.634 & 0.493 & 0.139 & 0.491 & 0.105 & 0.106 & 0.066 & 0.069 & \textbf{0.057}\\
	& max & 0.461 & 1.170 & 0.943 & 0.152 & 0.506 & 0.119 & 0.113 & 0.070 & 0.081 & \textbf{0.063} \\
 	& mean& 0.382 & 0.730 & 0.594 & 0.144 & 0.497 & 0.110 & 0.109 & 0.068 & 0.073	& \textbf{0.060} \\ 
 	& $\delta$ & 0.017	& 0.091	& 0.108	& 0.001	& 0.001	& 0.002	& 0.001	& 0.001	& 0.002	& \textbf{0.0008} \\ 
 	\hline
	\multirow{3}{*}
	{\!\!\!\!\! brain \!\!\! $(\!720 \! \times \! 720 \!)$ \!\!\!\!\!} 
 	& min & 1.61 & 0.66 &  0.66 & 0.96 & 0.49 & 0.38 & 0.26 & 0.17 &  0.17 & \textbf{0.13}\\
 	& max & 2.11 & 0.66 &  0.66 & 1.00 & 0.56 & 0.49 & 0.26 & 0.17 &  0.17 & \textbf{0.13}\\
 	& mean & 1.77 & 0.66 &  0.66 & 0.97 & 0.51 & 0.40 & 0.26 & 0.17 &  0.17 & \textbf{0.13}\\ 
 	& $\delta$ & 0.382 & 0.730 & 0.594 & 0.144 & 0.497 & 0.110 & 0.109 & 0.068 & 0.073	& \textbf{0.060} \\  	
 	\hline
	\multirow{3}{*}
	{\!\!\!\!\! fingerprint \!\!\! $(\!1024 \! \times \! 1024 \!)$ \!\!\!\!\!} 
 	& min & 1.565	 & 1.676 & 0.730 & 0.379 & 0.570 & 0.297 & 0.492 & 0.249 & 0.285 & \textbf{0.210}\\
 	& max & 1.640 & 2.310 & 1.100 & 0.465 & 0.703 & 0.370 & 0.612 & 0.316 & 0.362 & \textbf{0.259}\\
 	& mean & 1.584 & 1.793 & 0.797 & 0.393 & 0.583 & 0.307 & 0.504 & 0.256 & 0.293 & \textbf{0.217}\\
 	& $\delta$ & 0.015 & 0.151 & 0.086 & 0.018 & 0.030 & 0.0150 & 0.026 & 0.013 & 0.015 & \textbf{0.010} \\  	 
 	\hline
	\multirow{3}{*}
	{\!\!\!\!\! cartoon \!\!\! $(\!1024 \! \times \! 1024 \!)$ \!\!\!\!\!} 
 	& min & 2.011 & 1.746 & 0.901 & 0.373 & 0.511 & 0.308 & 0.434 & 0.237 & 0.263 & \textbf{0.200}\\
 	& max & 2.203 & 2.339 & 1.324 & 0.499 & 0.629 & 0.380 & 0.542 & 0.294 & 0.327 & \textbf{0.249}\\
 	& mean & 2.034 & 1.949 & 0.970 & 0.397 & 0.521 & 0.316 & 0.443 & 0.243 & 0.270 & \textbf{0.205}\\ 
 	& $\delta$ & 0.025 & 0.168 & 0.104 & 0.023 & 0.027 & 0.016 & 0.022 & 0.011 & 0.012 & \textbf{0.009} \\  	
 	\hline
	\multirow{3}{*}
	{\!\!\!\!\! texture \!\!\! $(\! 2048 \! \times \! 2048 \!)$ \!\!\!\!\!} 
 	& min & 7.016 & 3.608 & 1.685 & 0.656 & 0.596 & 0.676 & 0.488 & 0.447 & 0.453 & \textbf{0.438}\\
 	& max & 7.669 & 4.764 & 2.538 & 0.852 & 0.690 & 0.794 & 0.560 & 0.510 & 0.526 & \textbf{0.499}\\
 	& mean & 7.096 & 3.949 & 2.025 & 0.668 & 0.604 & 0.684 & 0.494 & 0.453 & 0.459 & \textbf{0.443}\\ 
 	& $\delta$ & 0.100 & 0.219 & 0.182 & 0.029 & 0.015 & 0.019 & 0.012 & 0.011 & 0.011 & \textbf{0.009} \\  	
 	\hline
	\multirow{3}{*}
	{\!\!\!\!\! artifact \!\!\! $(\!4096 \! \times \! 4096 \!)$ \!\!\!\!\!} 
 	& min & 125.198 & 21.620 & 6.439 & 5.186 & 5.089 & 4.099 & 3.632 & 3.280 & 3.268 & \textbf{3.097}\\
 	& max & 127.943 & 24.469 & 8.259 & 5.933 & 6.061 & 4.842 & 4.365 & 3.927 & 3.908 & \textbf{3.700}\\
 	& mean & 126.633 & 22.082 & 7.691 & 5.232 & 5.135 & 4.132 & 3.668 & 3.302 & 3.295 & \textbf{3.120}\\
 	& $\delta$ & 0.887 & 0.412 & 0.390 & 0.121 & 0.163 & 0.125 & 0.079 & 0.065 & 0.062 & \textbf{0.059} \\  	
 	\hline
	\end{tabular}
    \end{small}
\end{sidewaystable}

\begin{figure}[t]
\footnotesize
\centering
\includegraphics[width=0.90\linewidth]{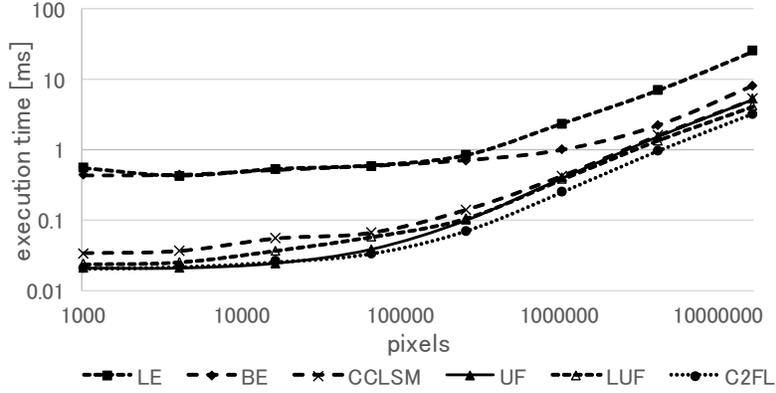}
\vskip 1mm
(a) Execution time versus image pixels
\vskip 2mm

\includegraphics[width=0.90\linewidth]{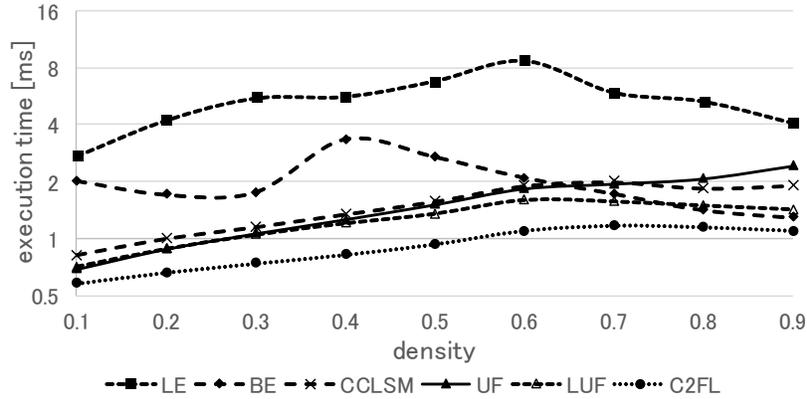}
\vskip 1mm
(b) Execution time versus label density
\caption{Execution time versus label density and image pixels}
\label{fig:execution-time-evaluation}
\end{figure}%

\begin{figure}[t]
\centering
\footnotesize
	\begin{minipage}[b]{0.31\linewidth}
 		\centering
 		\subfloat[Lena]
		{
 	 		\begin{overpic}[width=1\textwidth]
 	 			{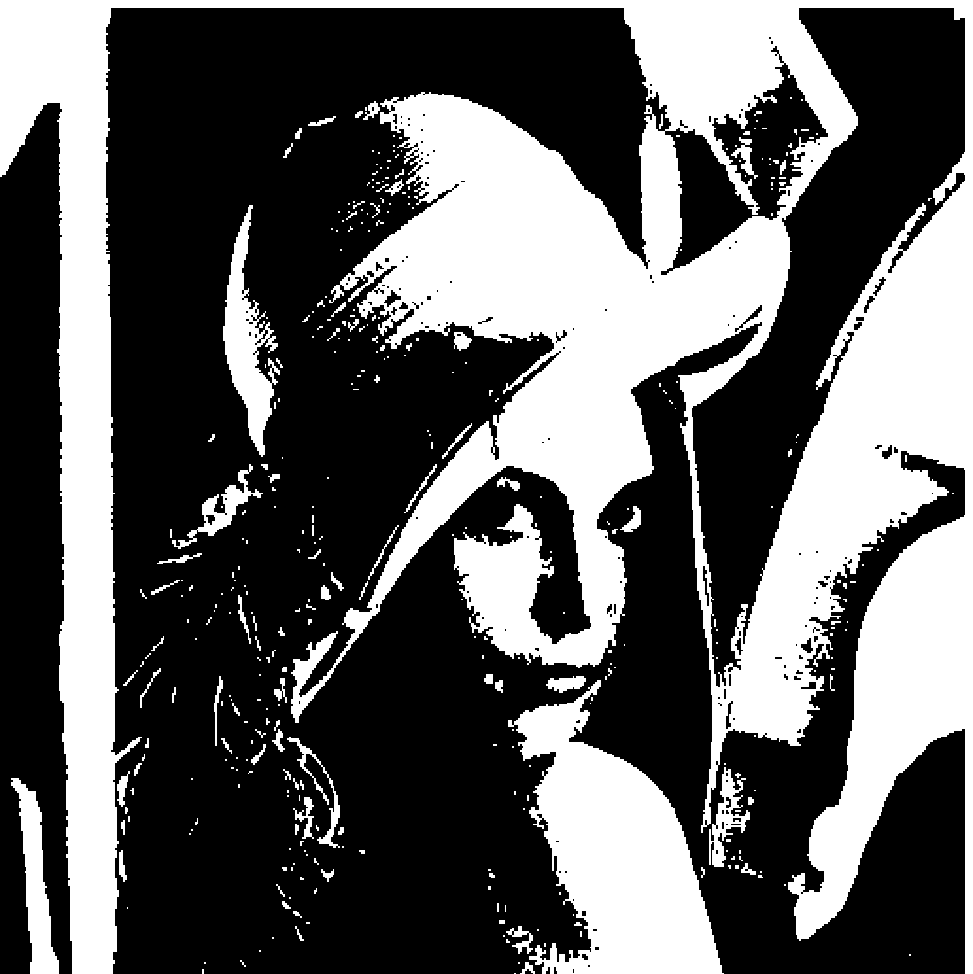}
 		\end{overpic}
 	 	}
	\end{minipage}
\hskip 1mm
	\begin{minipage}[b]{0.31\linewidth}
 		\centering
 		\subfloat[Brain]
		{
 			\begin{overpic}[width=1\textwidth]
 	 			{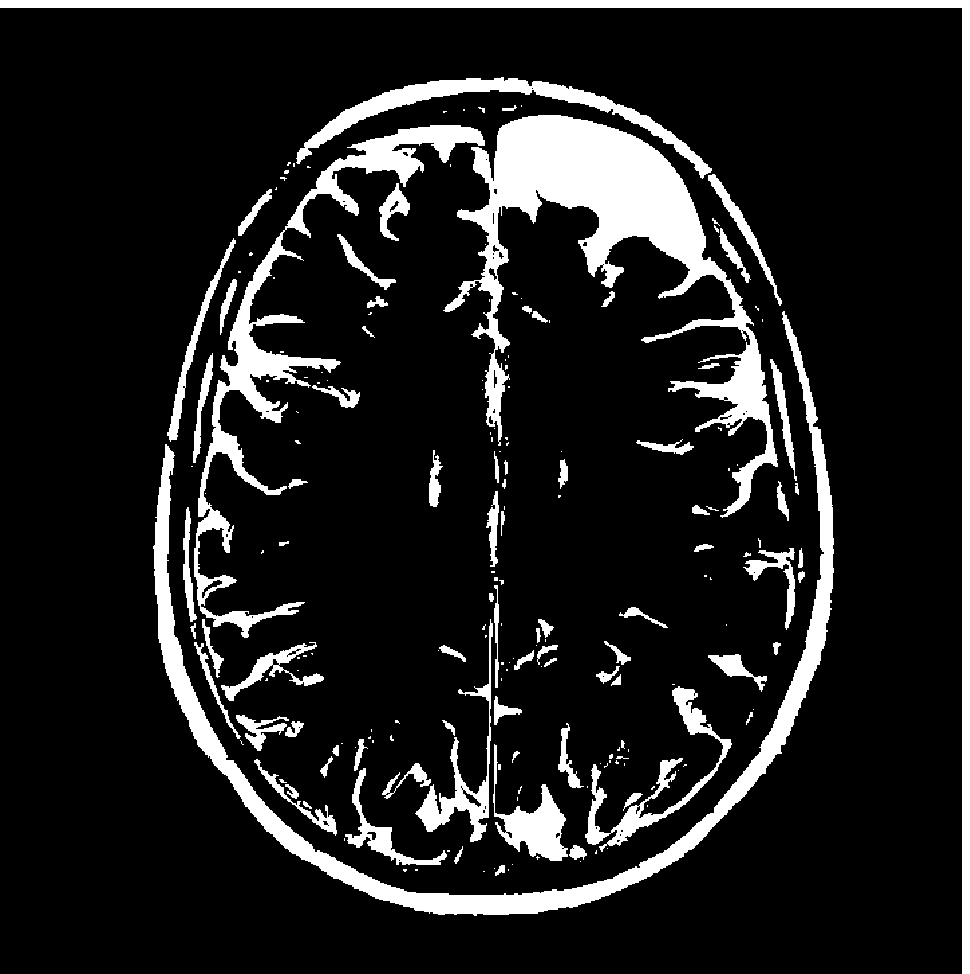}
 		\end{overpic}
 		}
	\end{minipage}
\hskip 1mm
	\begin{minipage}[b]{0.31\linewidth}
 		\centering
 		\subfloat[Fingerprint]
		{
 			\begin{overpic}[width=1\textwidth]
 	 			{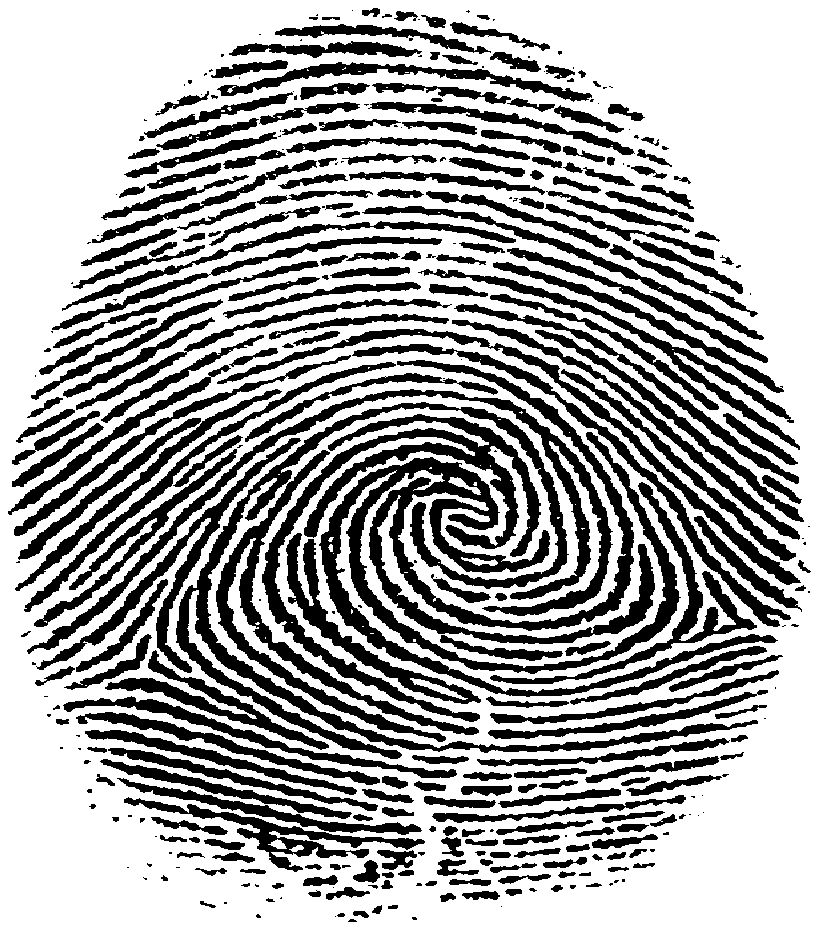}
 		\end{overpic}
 		}
	\end{minipage}

	\vskip 2mm

	\begin{minipage}[b]{0.31\linewidth}
 		\centering
 		\subfloat[Cartoon]
		{
 	 		\begin{overpic}[width=1\textwidth]
 	 			{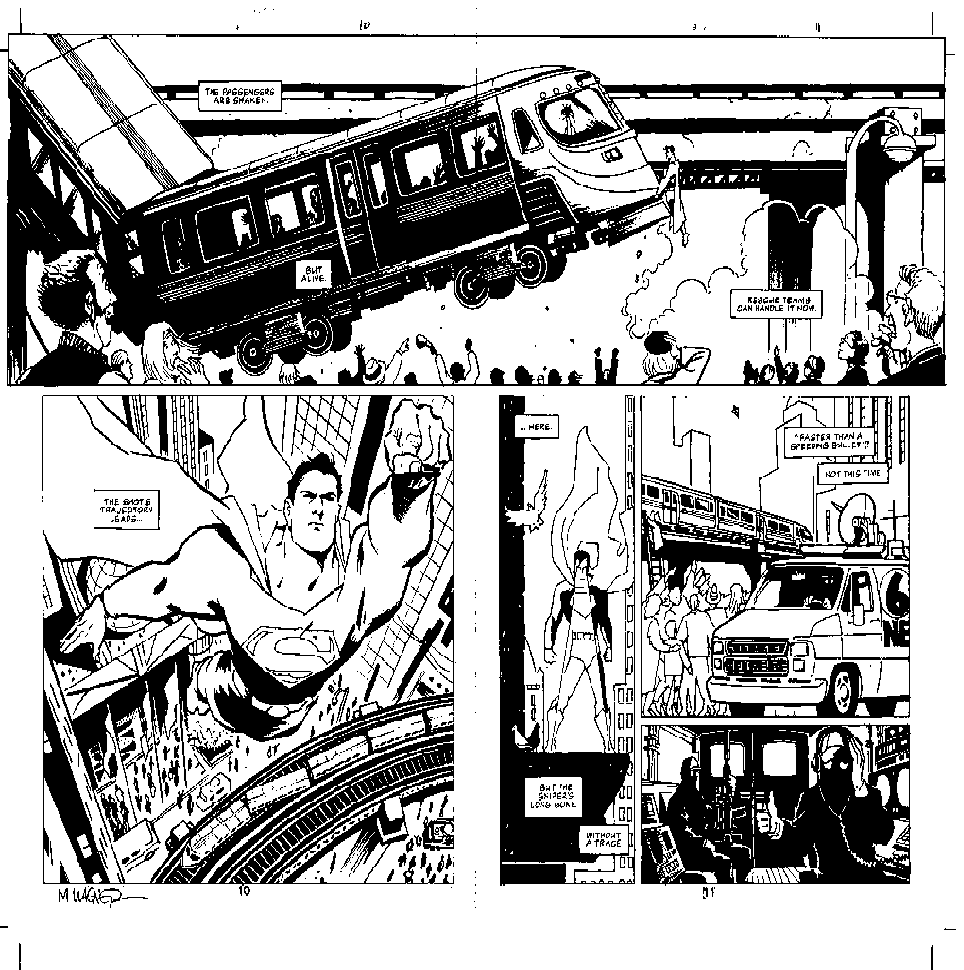}
 		\end{overpic}
 	 	}
	\end{minipage}
\hskip 1mm
	\begin{minipage}[b]{0.31\linewidth}
 		\centering
 		\subfloat[Texture]
		{
 			\begin{overpic}[width=1\textwidth]
 	 			{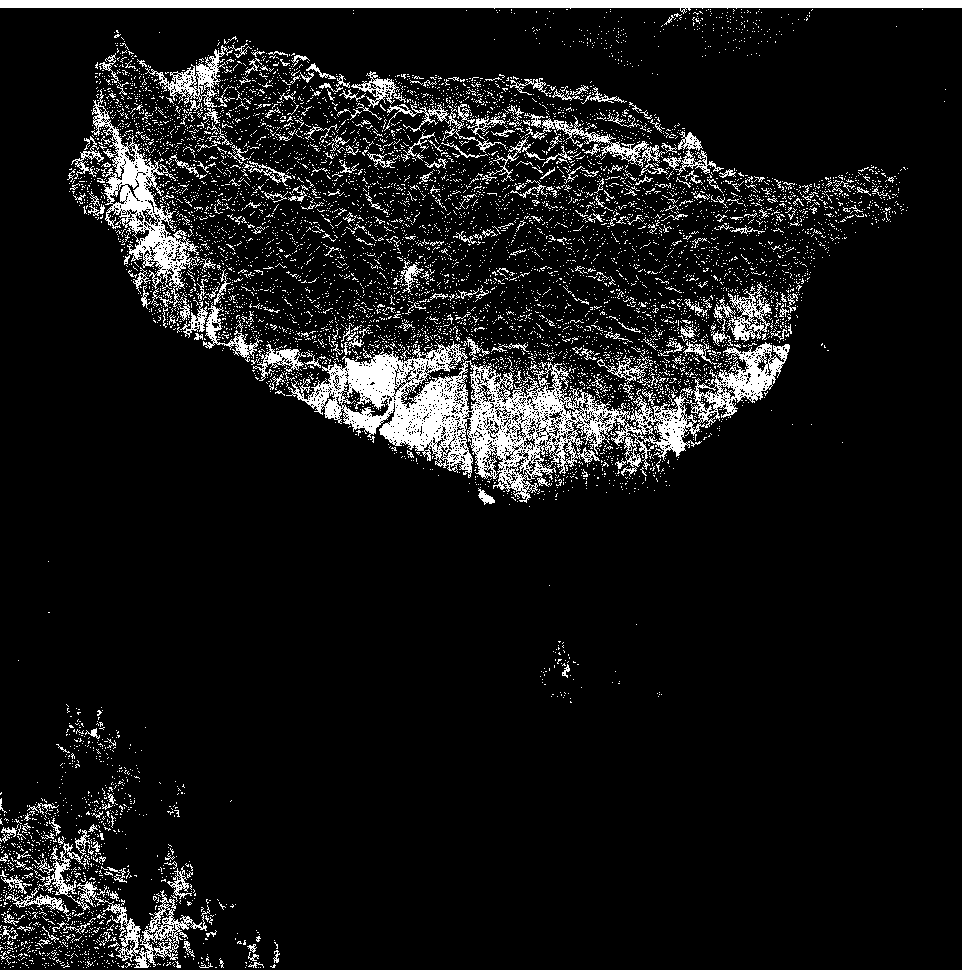}
 		\end{overpic}
 		}
	\end{minipage}
\hskip 1mm
	\begin{minipage}[b]{0.31\linewidth}
 		\centering
 		\subfloat[Artifact]
		{
 			\begin{overpic}[width=1\textwidth]
 	 			{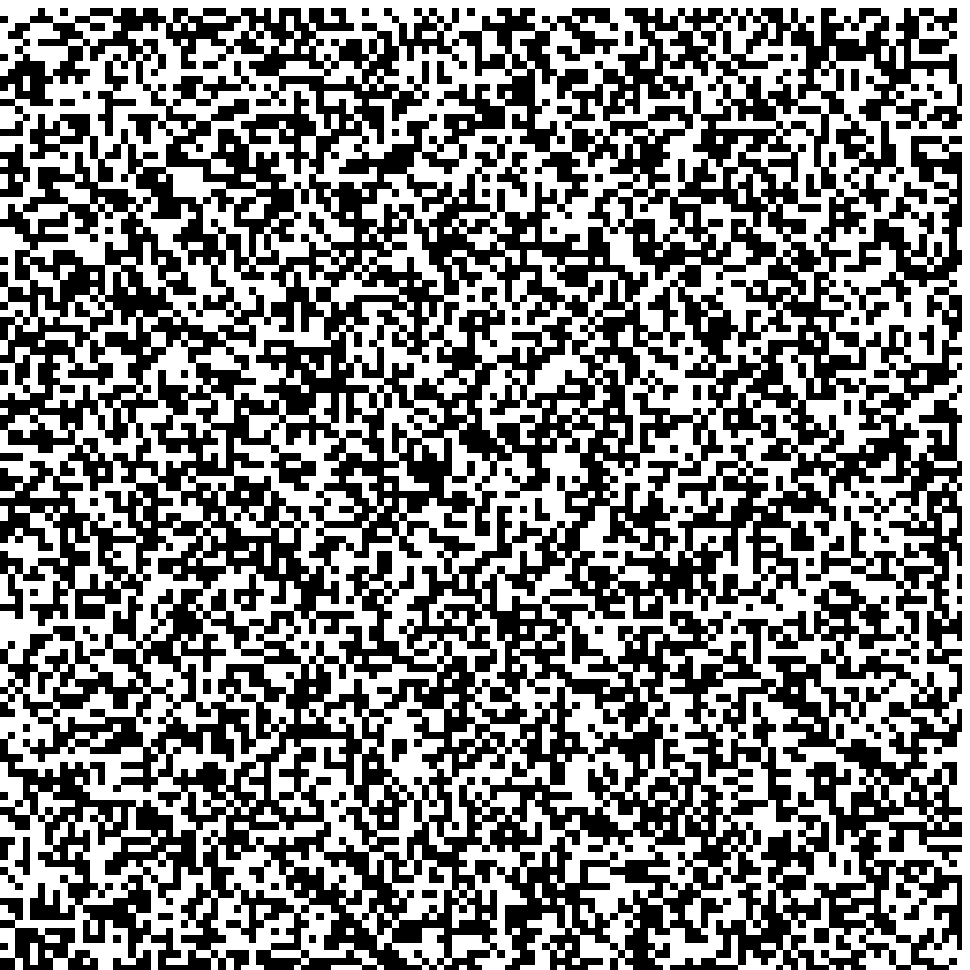}
 		\end{overpic}
 		}
	\end{minipage}	

\caption{Six representative images.}
\label{fig:six-representative-images}
\vskip -3mm
\end{figure}

\subsection{Comparisons with existing algorithms}

For the evaluation of execution time, the minimum, maximum, mean, and standard deviation, over 100 runs are collected to remove any fluctuations caused by the other tasks executed by the operation system.
All the algorithms were implemented based on four-neighbor connection, except BE and CPUCCL.
BE assumes that the pixels in a $2 \times 2$ block share the same label such that it is infeasible to perform with four-connectivity.
CPUCCL performs CCL using a pixel prediction strategy to avoid repeatedly scanning the same pixels multiple times. 
The code for its implementation is borrowed from \cite{grana2016yacclab}.

We first evaluate the algorithms using the same synthetic images.
Fig.~\ref{fig:execution-time-evaluation} (a) shows how the algorithms work with images of increasing size.
The execution time of all algorithms increases linearly with the expansion of input images.
Our method is proved to be scalable and able to outperform all the other methods.
Another experimental result, shown in Fig.~\ref{fig:execution-time-evaluation} (b), highlights the efficiency of the algorithms with images of various label densities.
It indicates that the computation is efficient regardless of the label density is low or high while the worst case appears around the middle densities.
Lower or higher densities present simple connections and consequently less computation, while the middle densities present complex connections.
It can be inferred that our approach has the best performance among all the densities.
It is able to label a $2048 \times 2048$ image with arbitrary density within $1.2~$ms

We also selected the six images shown in Fig.~\ref{fig:six-representative-images}, comprising a natural image, medical image, texture, and artifact image, to prove the performance of our proposed method.
The resolution and experimental results of the six images are listed in Tab.~\ref{tab:speedup}.
The comparison among NC2FL, RC2FL, CC2FL, and C2FL states the effectiveness of the coarse-to-fine strategy.
NC2FL always takes longer to complete one segmentation than the other three methods.
Meanwhile, C2FL is always more efficient than RC2FL and CC2FL.
The comparison among all the algorithms is evidence that our approach outperforms all the others for all the images.
It should also be noted that the standard deviation of the execution time of our approach is quite small for all the images, which demonstrates stable computation.

\section{Conclusion}

In this paper, we proposed a novel parallel approach with a coarse-to-fine strategy to accelerate the solution of CCL issues.
Our method first employs coarse labeling to reduce the complexity of a local block, and then applies a refinement to solve the local labeling.
In the block merge stage, we launch a low number of threads to analyze the connectivities along block boundary.
As a result, the proposed method is sufficiently capable of performing CCL with CUDA on GPU.
We evaluated the effectiveness of the coarse-to-fine strategy and compared it with existing GPU and CPU implementations.
Experimental results show that our method outperforms all existing parallel approaches between $29\%$ and $80\%$ on average.
Meanwhile, it proved that our method has good scalability in term of various image sizes and stability in terms of various label densities.
The reference code for the method is available at \url{https://github.com/sevenlovechen/parallel_CCL}. 


\bibliography{mybibfile}

\begin{thebibliography}{10}
\expandafter\ifx\csname url\endcsname\relax
  \def\url#1{\texttt{#1}}\fi
\expandafter\ifx\csname urlprefix\endcsname\relax\def\urlprefix{URL }\fi
\expandafter\ifx\csname href\endcsname\relax
  \def\href#1#2{#2} \def\path#1{#1}\fi

\bibitem{baxes1994digital}
G.~A. Baxes, Digital image processing: principles and applications, Wiley New
  York, 1994.

\bibitem{suzuki2003massive}
K.~Suzuki, S.~G. Armato, F.~Li, S.~Sone, et~al., Massive training artificial
  neural network (mtann) for reduction of false positives in computerized
  detection of lung nodules in low-dose computed tomography, Medical physics
  30~(7) (2003) 1602--1617.

\bibitem{suzuki2008mixture}
K.~Suzuki, H.~Yoshida, J.~N{\"a}ppi, S.~G. Armato, A.~H. Dachman, Mixture of
  expert 3d massive-training anns for reduction of multiple types of false
  positives in cad for detection of polyps in ct colonography, Medical physics
  35~(2) (2008) 694--703.

\bibitem{song2017motion}
W.~Song, D.~Wu, Y.~Xi, Y.~W. Park, K.~Cho, Motion-based skin region of interest
  detection with a real-time connected component labeling algorithm, Multimedia
  Tools and Applications 76~(9) (2017) 11199--11214.

\bibitem{chen2013fast}
J.~Chen, Q.~Gu, H.~Gao, T.~Aoyama, T.~Takaki, I.~Ishii, Fast 3-d shape
  measurement using blink-dot projection, in: Intelligent Robots and Systems
  (IROS), 2013 IEEE/RSJ International Conference on, IEEE, 2013, pp.
  2683--2688.

\bibitem{chen2015blink}
J.~Chen, Q.~Gu, T.~Aoyama, T.~Takaki, I.~Ishii, Blink-spot projection method
  for fast three-dimensional shape measurement, Journal of Robotics and
  Mechatronics 27~(4) (2015) 430--443.

\bibitem{guler2016real}
P.~Guler, D.~Emeksiz, A.~Temizel, M.~Teke, T.~T. Temizel, Real-time
  multi-camera video analytics system on gpu, Journal of Real-Time Image
  Processing 11~(3) (2016) 457--472.

\bibitem{he2017connected}
L.~He, X.~Ren, Q.~Gao, X.~Zhao, B.~Yao, Y.~Chao, The connected-component
  labeling problem: A review of state-of-the-art algorithms, Pattern
  Recognition 70 (2017) 25--43.

\bibitem{cabaret2014review}
L.~Cabaret, L.~Lacassagne, L.~Oudni, A review of world's fastest connected
  component labeling algorithms: Speed and energy estimation, in: Design and
  Architectures for Signal and Image Processing (DASIP), 2014 Conference on,
  IEEE, 2014, pp. 1--6.

\bibitem{he2011two}
L.~He, Y.~Chao, K.~Suzuki, Two efficient label-equivalence-based
  connected-component labeling algorithms for 3-d binary images, IEEE
  Transactions on Image Processing 20~(8) (2011) 2122--2134.

\bibitem{martin2007hybrid}
J.~Mart{\'\i}n-Herrero, Hybrid object labelling in digital images, Machine
  Vision and Applications 18~(1) (2007) 1--15.

\bibitem{chang2004linear}
F.~Chang, C.-J. Chen, C.-J. Lu, A linear-time component-labeling algorithm
  using contour tracing technique, computer vision and image understanding
  93~(2) (2004) 206--220.

\bibitem{he2009fast}
L.~He, Y.~Chao, K.~Suzuki, K.~Wu, Fast connected-component labeling, Pattern
  Recognition 42~(9) (2009) 1977--1987.

\bibitem{he2008run}
L.~He, Y.~Chao, K.~Suzuki, A run-based two-scan labeling algorithm, IEEE
  Transactions on Image Processing 17~(5) (2008) 749--756.

\bibitem{he2010efficient}
L.~He, Y.~Chao, K.~Suzuki, An efficient first-scan method for
  label-equivalence-based labeling algorithms, Pattern Recognition Letters
  31~(1) (2010) 28--35.

\bibitem{grana2010optimized}
C.~Grana, D.~Borghesani, R.~Cucchiara, Optimized block-based connected
  components labeling with decision trees, IEEE Transactions on Image
  Processing 19~(6) (2010) 1596--1609.

\bibitem{he2014configuration}
L.~He, X.~Zhao, Y.~Chao, K.~Suzuki, Configuration-transition-based
  connected-component labeling, IEEE Transactions on Image Processing 23~(2)
  (2014) 943--951.

\bibitem{johnston2008fpga}
C.~T. Johnston, D.~G. Bailey, Fpga implementation of a single pass connected
  components algorithm, in: Electronic Design, Test and Applications, 2008.
  DELTA 2008. 4th IEEE International Symposium on, IEEE, 2008, pp. 228--231.

\bibitem{gu2013fast}
Q.~Gu, T.~Takaki, I.~Ishii, Fast fpga-based multiobject feature extraction,
  IEEE Transactions on Circuits and Systems for Video Technology 23~(1) (2013)
  30--45.

\bibitem{manohar1989connected}
M.~Manohar, H.~Ramapriyan, Connected component labeling of binary images on a
  mesh connected massively parallel processor, Computer vision, graphics, and
  image processing 45~(2) (1989) 133--149.

\bibitem{dewar1987parallel}
R.~Dewar, C.~Harris, Parallel computation of cluster properties: application to
  2d percolation, Journal of Physics A: Mathematical and General 20~(4) (1987)
  985.

\bibitem{nickolls2008scalable}
J.~Nickolls, I.~Buck, M.~Garland, K.~Skadron, Scalable parallel programming
  with cuda, Queue 6~(2) (2008) 40--53.

\bibitem{sanders2010cuda}
J.~Sanders, E.~Kandrot, CUDA by Example: An Introduction to General-Purpose GPU
  Programming, Portable Documents, Addison-Wesley Professional, 2010.

\bibitem{jung2010parallel}
I.-Y. Jung, C.-S. Jeong, Parallel connected-component labeling algorithm for
  gpgpu applications, in: Communications and Information Technologies (ISCIT),
  2010 International Symposium on, IEEE, 2010, pp. 1149--1153.

\bibitem{hawick2010parallel}
K.~A. Hawick, A.~Leist, D.~P. Playne, Parallel graph component labelling with
  gpus and cuda, Parallel Computing 36~(12) (2010) 655--678.

\bibitem{kalentev2011connected}
O.~Kalentev, A.~Rai, S.~Kemnitz, R.~Schneider, Connected component labeling on
  a 2d grid using cuda, Journal of Parallel and Distributed Computing 71~(4)
  (2011) 615--620.

\bibitem{soh2014fast}
Y.~Soh, H.~Ashraf, Y.~Hae, I.~Kim, Fast parallel connected component labeling
  algorithms using cuda based on 8-directional label selection, Int. J. Latest
  Res. Sci. Technol 3~(2) (2014) 187--190.

\bibitem{zavalishin2016block}
S.~Zavalishin, I.~Safonov, Y.~Bekhtin, I.~Kurilin, Block equivalence algorithm
  for labeling 2d and 3d images on gpu, Electronic Imaging 2016~(2) (2016)
  1--7.

\bibitem{cormen2009introduction}
T.~H. Cormen, Introduction to algorithms, MIT press, 2009.

\bibitem{oliveira2010study}
V.~Oliveira, R.~Lotufo, A study on connected components labeling algorithms
  using gpus, in: SIBGRAPI, Vol.~3, 2010, p.~4.

\bibitem{stava2010connected}
O.~Stava, B.~Benes, Connected component labeling in cuda, Hwu., WW (Ed.), GPU
  Computing Gems.

\bibitem{kumar2016real}
P.~Kumar, A.~Singhal, S.~Mehta, A.~Mittal, Real-time moving object detection
  algorithm on high-resolution videos using gpus, Journal of Real-Time Image
  Processing 11~(1) (2016) 93--109.

\bibitem{park2000fast}
J.-M. Park, C.~G. Looney, H.-C. Chen, Fast connected component labeling
  algorithm using a divide and conquer technique., Computers and Their
  Applications 4 (2000) 4--7.

\bibitem{cormen1990floyd}
T.~H. Cormen, C.~E. Leiserson, R.~L. Rivest, The floyd-warshall algorithm,
  Introduction to Algorithms (1990) 558--565.

\bibitem{chen2011block}
P.~Chen, H.~Zhao, C.~Tao, H.~Sang, Block-run-based connected component
  labelling algorithm for gpgpu using shared memory, Electronics letters
  47~(24) (2011) 1309--1311.

\bibitem{zhao2010stripe}
H.~Zhao, Y.~Fan, T.~Zhang, H.~Sang, Stripe-based connected components
  labelling, Electronics letters 46~(21) (2010) 1434--1436.

\bibitem{paravecino2014accelerated}
F.~N. Paravecino, D.~Kaeli, Accelerated connected component labeling using cuda
  framework, in: International Conference on Computer Vision and Graphics,
  Springer, 2014, pp. 502--509.

\bibitem{xu2014graph}
Q.~Xu, H.~Jeon, M.~Annavaram, Graph processing on gpus: Where are the
  bottlenecks?, in: Workload Characterization (IISWC), 2014 IEEE International
  Symposium on, IEEE, 2014, pp. 140--149.

\bibitem{yonehara2015line}
K.~Yonehara, K.~Aizawa, A line-based connected component labeling algorithm
  using gpus, in: Computing and Networking (CANDAR), 2015 Third International
  Symposium on, IEEE, 2015, pp. 341--345.

\bibitem{nvidia2015toolkit}
C.~Nvidia, Toolkit documentation v7. 0, Nvidia Corporation.

\bibitem{grana2016optimized}
C.~Grana, L.~Baraldi, F.~Bolelli, Optimized connected components labeling with
  pixel prediction, in: International Conference on Advanced Concepts for
  Intelligent Vision Systems, Springer, 2016, pp. 431--440.

\bibitem{grana2016yacclab}
C.~Grana, F.~Bolelli, L.~Baraldi, R.~Vezzani, Yacclab-yet another connected
  components labeling benchmark, in: Pattern Recognition (ICPR), 2016 23rd
  International Conference on, IEEE, 2016, pp. 3109--3114.

\end{thebibliography}

\end{document}